\pdfoutput=1

\documentclass[11pt]{article}

\usepackage{acl}

\usepackage{times}
\usepackage{latexsym}

\usepackage[T1]{fontenc}

\usepackage[utf8]{inputenc}

\usepackage{microtype}

\usepackage{amsmath,amsfonts,bm}
\usepackage{hyperref}
\usepackage{url}

\usepackage{multirow}
\usepackage{ulem}
\usepackage{makecell}
\usepackage{subfig}
\usepackage[justification=centering,justification=justified]{caption}
\usepackage[pdftex]{graphicx}
\usepackage{arydshln}
\usepackage{xcolor}
\usepackage{booktabs}
\usepackage[leftcaption]{sidecap}

\usepackage[section]{placeins}

\newcommand{\vtot}{\texttt{vec2text}} 
\newcommand{\ttot}{\texttt{text2text}} 
\newcommand{\ttov}{\texttt{text2vec}} 

\newcommand*\samethanks[1][\value{\thefootnote}]{\footnotemark[#1]}

\title{\texttt{vec2text} with Round-Trip Translations}

\author{Geoffrey Cideron\quad\textbf{Sertan Girgin}\quad\textbf{Anton Raichuk}\\ \textbf{Olivier Pietquin}\quad\textbf{Olivier Bachem}\thanks{\quad Equal Supervision}\quad\textbf{L\'eonard Hussenot}\samethanks[1]\\
Google Research, Brain Team
}

\begin{document}

\maketitle

\begin{abstract}
  \looseness=-1
  We investigate models that can generate arbitrary natural language text (e.g. all English sentences) from a bounded, convex and well-behaved control space. We call them \textit{universal} \vtot{} \textit{models}. Such models would allow making semantic decisions in the vector space (e.g. via reinforcement learning) while the natural language generation is handled by the  \vtot{} model. We propose four desired properties -- \textit{universality}, \textit{diversity}, \textit{fluency}, and \textit{semantic structure} -- that such  \vtot{} models should possess and we provide quantitative and qualitative methods to assess them. We implement a  \vtot{} model by adding a bottleneck to a 250M parameters Transformer model and training it with an auto-encoding objective on 400M sentences (10B tokens) extracted from a massive web corpus. We propose a simple data augmentation technique based on round-trip translations and show in extensive experiments that the resulting  \vtot{} model surprisingly leads to vector spaces that fulfill our four desired properties and that this model strongly outperforms both standard and denoising auto-encoders.%

\end{abstract}

\section{Introduction}
\looseness=-1
In recent years, large language models (LLMs) have steadily improved in performance with impressive results in a large number of natural language understanding and generation tasks.
LLMs are commonly used in two specific ways:
First, models such as BERT \citep{devlin2018bert} have been used to obtain vector representations for input text (we call this the \texttt{text2vec} setting) which are then often used for natural language understanding tasks such as natural language inference, sentiment analysis, and paraphrase detection.
Second, models such as T5 \citep{raffel2019exploring} and GPT \citep{radford2019language,brown2020language} have been used to generate output text based on some input text (this is commonly denoted by the \texttt{text2text} setting). Common applications for such models are for example machine translation \citep{raffel2019exploring,vaswani2017attention,radford2019language,brown2020language,lewis2019bart,song2019mass} and summarization \citep{raffel2019exploring,radford2019language,lewis2019bart,dong2019unified,edunov-etal-2019-pre,song2019mass}.

\looseness=-1
In contrast to the \texttt{text2vec} and \texttt{text2text} settings described above, we focus in this paper on the \vtot{} setting: 
We explore models that take as input fixed size vectors and generate natural language text.
This setting investigates a fundamental research question: 
Is it possible to use continuous vector spaces to control the output of LLMs?
Such controllability would allow decoupling natural language generation from decision making. 
Then, reinforcement learning algorithms \citep{sutton2018reinforcement} could be used to control the semantics in a continuous actions space and a \vtot{} model could turn the taken action into natural text.

While there has been a variety of prior work on \vtot{} models \citep{bowman2015generating,pmlr-v119-shen20c,cifka2018eval,hu2017toward,zhao2018adversarially,li-etal-2020-optimus}, the majority of such work has focused on training relatively small models on relatively small and specialized datasets (e.g. the Yelp dataset with 500K sentences with less than 16 words \citep{pmlr-v119-shen20c}, Yahoo answers with 500K sentences \citep{yang2017improved}) for a limited set of downstream tasks such as style transfer
\citep{zhao2018adversarially,shen2017style,mai2020plug,hu2017toward,li-etal-2020-optimus}.
We take a different approach than prior work and investigate whether it is possible to train what we call universal \vtot{} models which intuitively should be able to generate arbitrary text,  not just for one specific task.
Furthermore, we argue that such a universal \vtot{} model should have a "nicely" structured input space that provides semantic controllability over the generated text.
In this paper, we focus on investigating such a universal \vtot{} model for the task of generating arbitrary English sentences. There are three key contributions in this paper:
\looseness=-1
First, we define the \vtot{} setting and propose four properties that such a model should possess:  universality, fluency, semantic structure, and diversity. We further derive several quantitative and qualitative analyses to assess a \vtot{} model in these dimensions.

Second, we implement and train a T5-based auto-encoder model on sentences extracted from the massive C4 dataset \citep{raffel2019exploring} and confirm commonly held beliefs that the decoder of such models have a poorly structured input space. Similarly, we implement the denoising approach from \citet{pmlr-v119-shen20c} and find that, while performing better than a standard auto-encoder, the input space still exhibits several deficiencies.

Third, we propose a novel approach that uses round trip translations (RTT) to obtain a nicely behaved \vtot{} model. Using an English-German and a German-English machine translation model, we translate all sentences in the C4 dataset to German and back to English (Table~\ref{rtt-table}) . We then train a T5-based auto-encoder model equipped with a bottleneck to predict the original sentences from the round-trip translated sentences. In extensive experiments, we show perhaps surprisingly that in contrast to the prior work this simple approach is sufficient to achieve all four properties.

\section{\vtot{} models and their desired properties }
\label{sec:properties}

\looseness=-1
\paragraph{\vtot{} decoders.} 
In this paper, we focus on the \vtot{} problem where the goal is to generate sentences from a convex and well-behaved vector space (see Figure~\ref{3figs} for a comparison to the \texttt{text2text} and \texttt{text2vec} settings). 
For the sake of simplicity, we focus on generating arbitrary English sentences but the setting and approach applies to the generation of arbitrary sequences.
More formally, let $\mathcal{X}$ be the set of all sequences up to a maximum length and $\mathcal{P}(\mathcal{X})$ the space of probability distributions over $\mathcal{X}$.
We define \vtot{} models (or equivalently \vtot{} decoders) to be of the form $D : \mathbb{R}^d \rightarrow \mathcal{P}(\mathcal{X}).$ 
Importantly, such \vtot{} decoders are assumed to be stochastic:
Given an input vector, they define a distribution over sequences, often in the form of a conditional language model.
\paragraph{Auxiliary text encoders.}
\looseness=-1
We will often also have access to auxiliary  encoders of the form $E : \mathcal{X} \rightarrow \mathbb{R}^d$
which embed text into fixed size vector embeddings.
Such encoders are also encountered in the \texttt{text2vec} settings where the goal is to learn good features for downstream tasks \citep{conneau2018senteval}.
The key difference is that we are interested in the generative capabilities of the \vtot{} decoder $D$ while the auxiliary encoder $E$ only provides a way to choose vector inputs to the decoder $D$.

\paragraph{Existing evaluation methods.}

Previous approaches proposed a variety of evaluations for \vtot{} models. However, they lack a formalization of the desired properties for universal \vtot{} models and the evaluations to assess these properties. In previous work, downstream tasks such as sentence classification \citep{montero-etal-2021-sentence} or language modeling \citep{bowman2015generating} were used to assess the quality of the learned models. The downside of this type of evaluation is that the performances on these downstream tasks -especially when some finetuning is done before the evaluation- are not straightforward indicators of the quality of universal \vtot{} models. Other evaluations directly assess the effect of latent space traversal (e.g. interpolation) on decoded sentences \citep{pmlr-v119-shen20c} but they lack a comprehensive analysis of the desired properties for such models and how their evaluation relate to them.  

\subsection{Desired properties of \vtot{} models}

To better understand what properties a \vtot{} model should possess we need to first have a closer look at how they work. They are essentially functions from a $d$-dimensional input space to the space of distributions over sentences. The first thing to consider is for which inputs the model should be well behaved. The naive approach would be to enforce this for arbitrary input vectors. However, this would mean that they have to work well on an unbounded input space of infinite volume which may be hard.

In this paper, we advocate for the following requirement: A ``good'' \vtot{} model should possess a bounded and convex \textit{control space} $C \subset \mathbb{R}^d$ where the model is well-behaved. The intuition is that this control space $C$ is where we actually want to use the model. The convexness is important as it guarantees that any point between two points in $C$ is also in $C$.

We propose four common sense properties that \vtot{} models should possess: \textit{universality}, \textit{diversity}, \textit{fluency}, and \textit{semantic structure}. 

\textbf{Universality.} For each English sentence, there should be an embedding in the control space $C$ that generates sentences that have the same meaning as the initial sentence with high probability. Intuitively, this property guarantees that any meaning can be expressed through a suitable choice of input to the \vtot{} model.

\textbf{Diversity.} Decoding from each embedding in the compact space C should have high entropy. Intuitively, this encourages that the \vtot{} model can express any meaning in a variety of different ways.

\textbf{Fluency.} Decoding from each embedding in the compact space C should lead to valid English sentences. Intuitively, this guarantees that there are no "holes" in $C$ where the \vtot{} model produces sentences that are not perceived as natural by humans.

\textbf{Semantic structure.} Two embeddings in the compact space C that are closeby should lead to similar distributions over sentences. Intuitively, this guarantees that the changes in meaning when the input to the \vtot{} model is changed is not abrupt.

\subsection{Qualitative assessment of \vtot{} models}

Our qualitative evaluations of \vtot{} models on the properties described in the previous section are made on decoded sentences generated from (i) known sentence embeddings, (ii) interpolated embeddings i.e. weighted average of the embedding of two known sentences, (iii) topic convex hull embeddings i.e. embeddings sampled from the convex hull of a set of embedded sentences from the same topic (e.g. weather, football). 

The \textbf{universality} property is qualitatively assessed by checking if the sentences generated from (i) have the same meaning as the known sentences. The \textbf{diversity} property is qualitatively assessed with (ii). Here, we examine the meaning consistency (e.g. rephrasing) between the decoded sentences. The \textbf{fluency} property is qualitatively assessed with (ii) as we examine if the generated sentences are plausible and of reasonable length. The \textbf{semantic structure} property is qualitatively assessed with (ii) and (iii): for the interpolated embeddings (ii), we examine if the generated sentences exhibit a semantic smoothness when varying the interpolation weights; for the topic convex hull embeddings (iii), we examine if the generated sentences are on the same topic as the sentences used to construct the convex hull.

\subsection{Quantitative assessment of \vtot{} models}

Our quantitative evaluations of \vtot{} models on the properties described in the previous section are made on two spaces: the interpolation space $\mathcal{I}$ which contains a set of embeddings from valid English sentences and some interpolation between them, and the noise space $\mathcal{O}$ which contains a set of embeddings from valid English sentences and noise perturbed versions of these embeddings. More details can be found in Section \ref{app:evaluation-details}.

The \textbf{universality} property is tested with the reconstruction accuracy of sentences unseen during training. For the \textbf{diversity} property, we computed the entropy and the entropy per token of the sentences generated from embeddings in $\mathcal{I}$ and $\mathcal{O}$. For the \textbf{fluency} property, we computed (1) the number of tokens (\# Tokens) and the max word repeat (Max Word Repeat), (2) the likelihood (LM LLH) and the likelihood per token (LM LLH / token) computed with an off-the-shelf language model. For the \textbf{diversity} property,  we computed the approximate Jeffreys divergence $J$\footnote{$J(p|q) = \frac{KL(p|q) + KL(q|p)}{2}$ with KL the Kullback-Leibler divergence, p and q two distributions.} between an anchor embedding and an embedding that progressively stray away from the anchor embedding by increasing the noise level or by decreasing the weight on the anchor sentence in the interpolation setting.

\section{Method}
In this section, we propose to train a T5-based universal sentence auto-encoder (Section~\ref{sec:universal-ae}) and how show to train it in order to obtain a good \vtot{} model (Section~\ref{sec:rtt-ae}).

\subsection{T5-based Universal Sentence Auto-Encoder}
\label{sec:universal-ae}
The simplest way to learn a \vtot{} model is to use an auto-encoder. It is composed of two components learned jointly: an encoder (a \ttov{} model)  $E : \mathcal{X} \rightarrow \mathbb{R}^d$ that maps the input sentence into a fixed-length vector and an auto regressive probabilistic decoder (a \vtot{} model) $D: \mathbb{R}^d \rightarrow \mathcal{P}(\mathcal{X})$ that maps a fixed-length vector to a distribution of sentences. The auto-encoder is trained by minimizing a cross entropy loss for input reconstruction:  $\displaystyle \min_\theta$ $\sum_{i=1}^{L} p_\theta(t_i| e, t_{<i})$ with $e$ the embedding of the sentence (the output of the encoder), $(t_1,..,t_L)$ the token of the input sentence ($t_0=\{\}$), and $\theta$ the parameters of the encoder and the decoder. This training procedure encourages the universality property while the fixed length sentence representation space gives a compact vector spaces which helps with the controllability. 
\looseness=-1
We propose to train a T5-based auto-encoder model \citep{raffel2019exploring} equipped with a bottleneck on sentences extracted from the massive C4 data \citep{raffel2019exploring} and use the decoder part as a \vtot{} model.
We use a pre-trained T5 model, as it was shown to be a general-purpose model that can achieve state-of-the-art performances on many downstream tasks.
T5 was previously used to learn sentence representation in \citet{ni2021sentence} where they focus on having a well structure sentence embedding by introducing a contrastive loss to pull together similar sentences and push them away from the negatives. However, \citet{ni2021sentence} don't learn a decoder (i.e. a \vtot{} model) which makes it impossible for them to generate sentences from the embedding space. \citet{park-lee-2021-finetuning} use T5 with a text variational auto-encoders \citep{bowman2015generating}. Their experiments mainly revolve around approaches to evaluate and mitigate posterior collapse and do not focus on how the model could be used as a \vtot{} model.

In our approach, we add a bottleneck to T5 in between the encoder and the decoder. The bottleneck is composed of two simple operations: i) mean over the token axis to create a fixed-length vector followed with ii) a linear projection. Hence, in our setting the \ttov{} model $E$ which outputs single-token sequences (i.e. sentence embeddings) corresponds to the T5 encoder followed by the bottleneck  while the \vtot{} model $D$ corresponds to the T5 decoder.

To create an universal \vtot{}  model, we rely on the massive C4 dataset \citep{raffel2019exploring} (750GB of text) that contains text scrapped from the web. To learn our sentence auto-encoder, we splitted the text in C4 into sentences to create the C4 sentence dataset (C4S)  which contains several billions of sentences. C4S combines two key properties: it is massive (billions of sentences) and it contains a wide variety of meaning as a wide variety of sites are scrapped. Hence, we call universal an auto-encoder trained on C4S with a good reconstruction accuracy.

\subsection{Round-Trip Translation for Auto-encoders}%
\label{sec:rtt-ae}
\looseness=-1
In order to train a  T5-based universal sentence auto-encoder, we first implemented the simplest approach that we call Vanilla AE where the input sentences match the target sentences. Previous works \citep{pmlr-v119-shen20c,bowman2015generating} showed that Vanilla AE is not able to produce a \vtot{} model with satisfying properties. As a second baseline, we implemented an approach inspired from \citet{pmlr-v119-shen20c} that we call AE + Denoising which uses denoising \citep{vincent2010stacked} to mitigate the lack of structure of the vanilla approach \citep{pmlr-v119-shen20c,montero-etal-2021-sentence}. To do so, we corrupt the original C4S dataset by dropping words (i.e. a sequence of token in between two white space tokens) in the sentences with a fixed probability of 20\%. AE + Denoising is learned by predicting the original sentence from its corrupted version as illustrated in the second row of Table~\ref{dataset-table}.

As explained by \citet{pmlr-v119-shen20c}, the intuition behind using denoising with auto-encoders is that the noise constraints the auto-encoder to put similar sentences (in terms of the denoising objective) next to each other in the latent space. However, the problem with denoising is that it maps together sentences that are close in edit distance but may have completely different meanings. \\ 
Ideally, we want to map semantically similar sentences next to each other. To achieve that, we investigate the use of paraphrasing in lieu of denoising. An automatic and scalable approach to create a dataset that contains the sentences and their paraphrasing is round-trip translation (RTT). As illustrated in Table~\ref{rtt-table}, a sentence in a source language (e.g. English) is translated to a pivot language (e.g. German); the translated sentence is then translated back to the source language. We relied on the publicly available English to German translation model and the German to English translation model that won the WMT21 competition \citep{tran2021facebook, akhbardeh2021findings} to create paraphrasing of the sentences in C4S as illustrated in Table~\ref{dataset-table}. We chose German for the pivot language as it produces more word reordering variations and the translation quality is good \citep{zhang2019paws}. The translations are decoded with nucleus sampling \citep{holtzman2019curious} with $p=0.9$ as we found it to produce diverse paraphrasing. We trained an auto-encoder that we called AE + RTT where the model is tasked to predict a sentence from its paraphrasing. We chose to predict the original sentence (as done for denoising) and not the rephrased one as it avoids potential side effects like translationese.

\section{Experimental results}

In this section, we evaluate the different models against the metrics defined in Section~\ref{sec:properties} that test the four desired properties: \textit{universality} (Section~\ref{sec:universality}), \textit{diversity} (Section~\ref{sec:diversity}), \textit{fluency} (Section~\ref{sec:no-holes}), and \textit{semantic structure} (Section~\ref{sec:semantic-structure}).
\subsection{Universality}
\label{sec:universality}
In this section, we evaluate the \textit{universality} property of our models. To do so, we compute the reconstruction accuracy of unseen sentences for two datasets: the evaluation dataset of C4S (Eval C4S) and the paraphrasing of the sentences in the evaluation dataset of C4S (Eval C4S RTT). The accuracy is computed per token with teacher forcing.

\begin{figure}[h]
\includegraphics[width=\columnwidth]{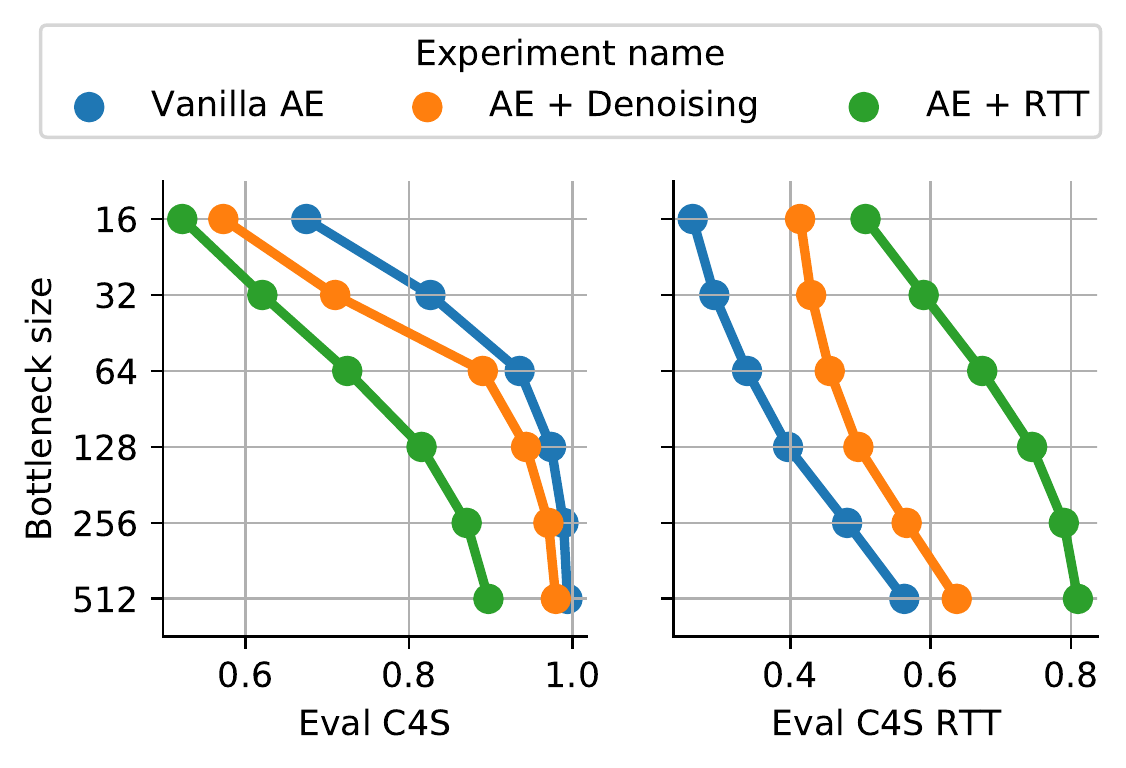}
\caption{Reconstruction accuracy on Eval C4S (left) and Eval C4S RTT (right).}
\label{fig:accuracy_tasks}
\end{figure}

Figure~\ref{fig:accuracy_tasks} shows a clear tradeoff between bottleneck size and accuracy. For bottleneck sizes 128 and higher, the accuracy for the Vanilla AE model on Eval C4S (Figure~\ref{fig:accuracy_tasks} left) is higher than 95\%. As expected, Vanilla AE has a better score on Eval C4S than the others as it is the only model trained for perfect reconstruction. AE + RTT still retains an accuracy above 80\% for bottleneck sizes equal or higher than 128. On Eval C4S RTT (Figure~\ref{fig:accuracy_tasks} right), the model trained with RTT performs best with more than 70\% of accuracy for bottleneck size equal or higher than 128. Unsurprisingly, the accuracy of Vanilla AE and AE + Denoising are low on Eval C4S RTT compared to Eval C4S e.g. for a bottleneck size of 128, Vanilla AE went from 95\% of accuracy on Eval C4S to less than 40\% of accuracy on Eval C4S RTT. 

The qualitative example in Section~\ref{app:universality-bottleneck-size} shows for the AE + RTT model the impact of the bottleneck size on the accuracy: the sentence is not well reconstructed for bottleneck sizes lower than 64 while for larger bottleneck sizes (64 and 128), the sentence is well reconstructed (either perfectly or with rephrasing). %
Section~\ref{app:universality-various-models} shows an example of sentence reconstruction for every model with a bottleneck size 128. Vanilla AE and AE + Denoising output the input sentence while AE + RTT rephrases the input sentence in various ways. In Table~\ref{decoding-sentence-ae-rtt}, AE + RTT rephrases the input sentence \textit{Why do you keep arguing that this is the case?} into \textit{Why do you keep arguing that this is \textbf{so}?} and \textit{Why do you \textbf{keep making claims} that this is the case?}.

\paragraph{Takeaway} Clear trade-off between the bottleneck size and reconstruction accuracy. The \textit{universality} property can be nearly satisfied for all the models (e.g. for a bottleneck size as low as 128 all the models have reconstruction accuracy around or above 80\%).

\subsection{Diversity}
\label{sec:diversity}
\begin{figure}[h]
\centering
\includegraphics[width=\columnwidth]{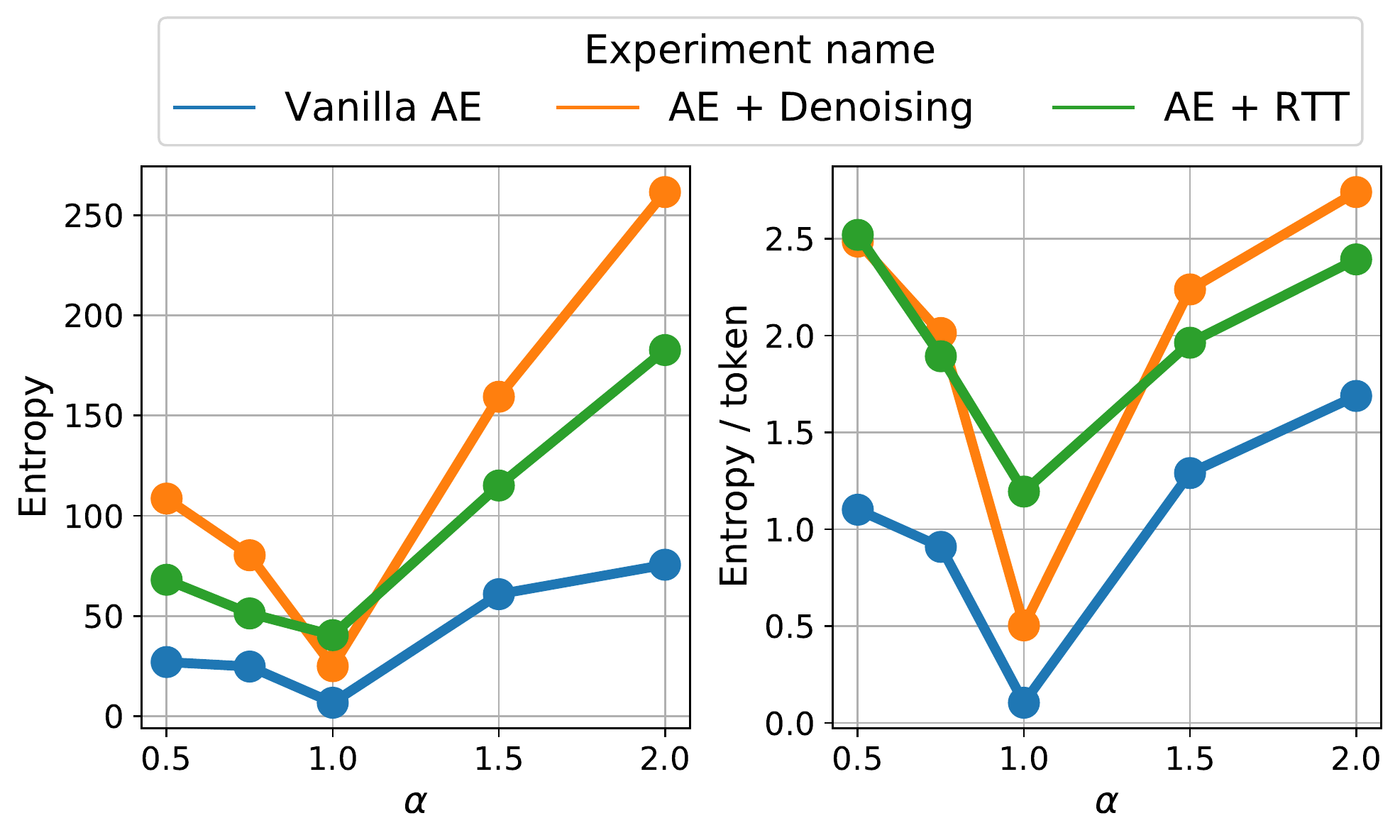}
\includegraphics[width=\columnwidth]{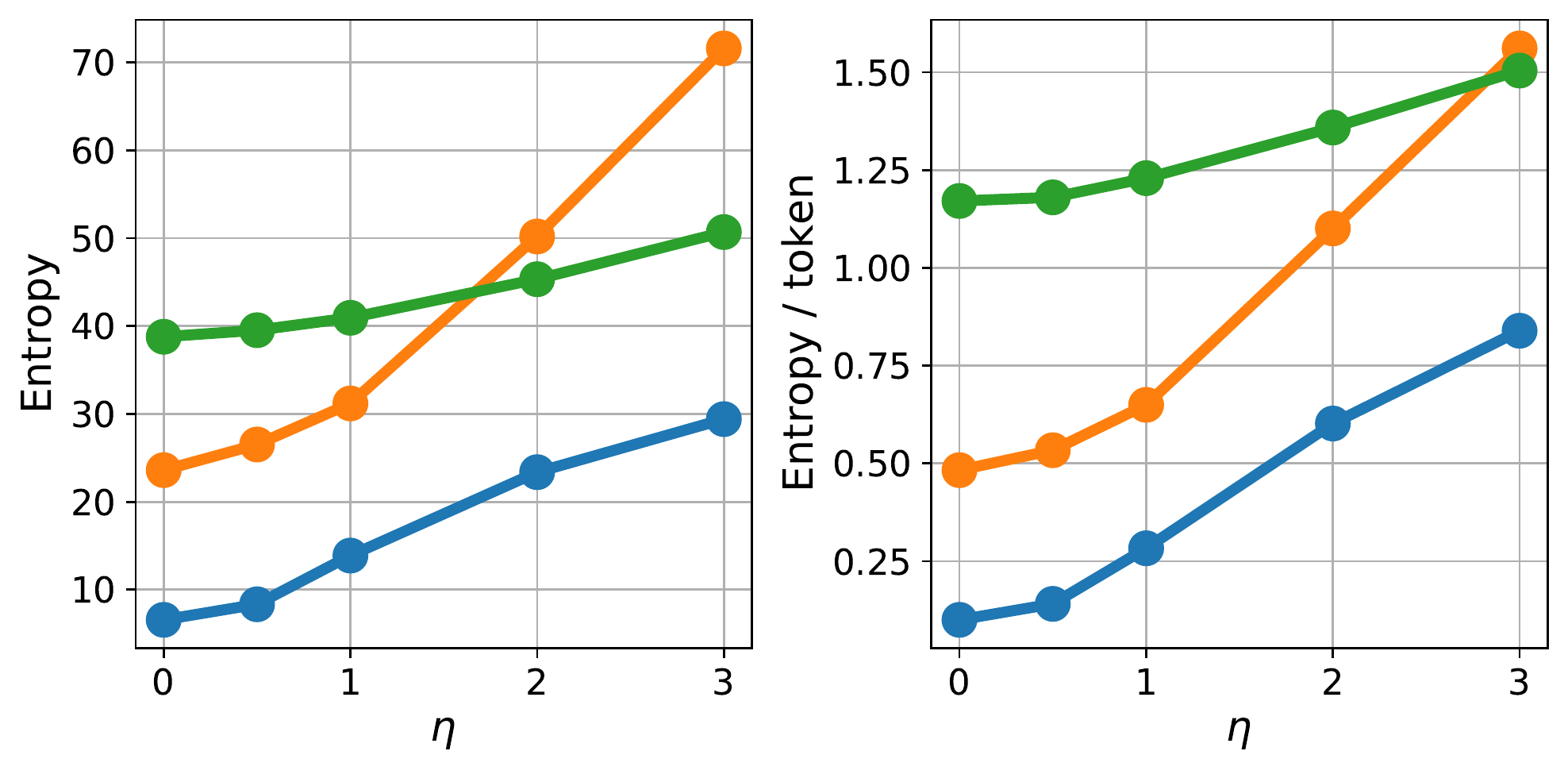}
\caption{Entropy and entropy per token of the distribution of decoded sentences for a bottleneck size 128. Decoded embeddings from interpolation (top row) or noise (bottom row).}
\label{fig:interpolation-noise-alpha-diversity}
\end{figure}

This section focuses on the diversity of the decoded sentences. The results are presented in Figure~\ref{fig:interpolation-noise-alpha-diversity} for a bottleneck size 128 while the full results are in Figure~\ref{fig:alpha-interpolation-length} (f)-(g) for the results on $\mathcal{I}$, and in Figure~\ref{fig:alpha-noise-length} (f)-(g) for the results on $\mathcal{O}$.

Figure~\ref{fig:interpolation-noise-alpha-diversity} shows that when there is no interpolation ($\alpha=1$) or no noise ($\eta=0$), AE + RTT has the highest entropy and entropy per token. This observation is consistent for bottleneck sizes above 64 (Figure~\ref{fig:alpha-interpolation-length} (f)-(g)). This relates with the observations made in the universality property section where the qualitative examples showed that the input sentences are rephrased in various ways for the models with RTT while the other models mostly reconstruct their inputs.
 
As shown in Figure~\ref{fig:interpolation-noise-alpha-diversity}, both RTT and denoising improve the entropy of the distribution of the decoded sentences when the embedding is computed from interpolation ($\alpha \neq 1$) or noise ($\eta > 0$). However, for the denoising perturbation, the surge in entropy is due to hallucinations which is a problem observed in natural language generation tasks \citep{ji2022survey,khandelwal2019sample}. This is shown in the second row of Table~\ref{interpolation-ae-denoising} where the decoded sentences are lengthy and lack consistency between each other in the interpolation setting: despite some words in common like \textit{children}, the generated sentences contains lots of details that are not consistent between each others. E.g. a sentence contains \textit{the reddish light was singing on the weather when you were on a bike} while another contains \textit{looking for a good song while watching the TV or the sleeping girl}.
The second row of Table~\ref{interpolation-ae-rtt} shows that the model using RTT suffers less from hallucinations as the produced sentences are of reasonable length and have better consistency. In the interpolation setting: out of the 5 decoded sentences two are about \textit{kids}/\textit{children} \textit{playing} and three are about \textit{kids}/\textit{children} \textit{watching}.

\textbf{Takeaway} AE + RTT is able to rephrase the same sentence in various ways and/or to produce sentences with consistency between each other. AE + Denoising either reconstructs the input like Vanilla AE or suffers from hallucinations that are characterized by lengthy sentences without consistency between each other.

\subsection{Fluency}
\label{sec:no-holes}
In this section, we investigate the \textit{fluency} property of our models on decoded sentences from interpolated or noisy embeddings. 
\begin{figure*}[h]
    \centering
    \includegraphics[width=\linewidth]{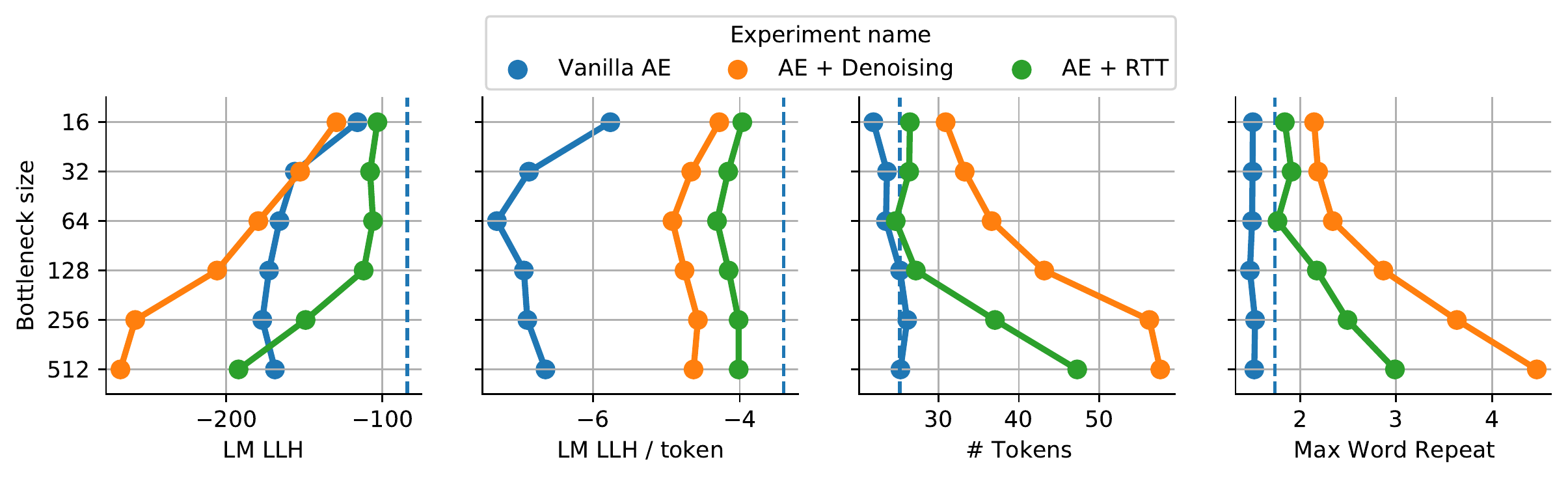}
    \caption{Fluency metrics for various bottleneck sizes on sentences decoded from interpolated embeddings $z = 0.5 \times z_1 + 0.5 \times z_2$ with $z_1$ and $z_2$ the embedding of two sentences (from the evaluation dataset of C4S). The dotted lines correspond to the statistics computed on 100 000 sentences from C4S.}
    \label{fig:interpolation}
\end{figure*}

\begin{figure*}[h]
\centering
\subfloat{%
  \includegraphics[clip,width=\linewidth]{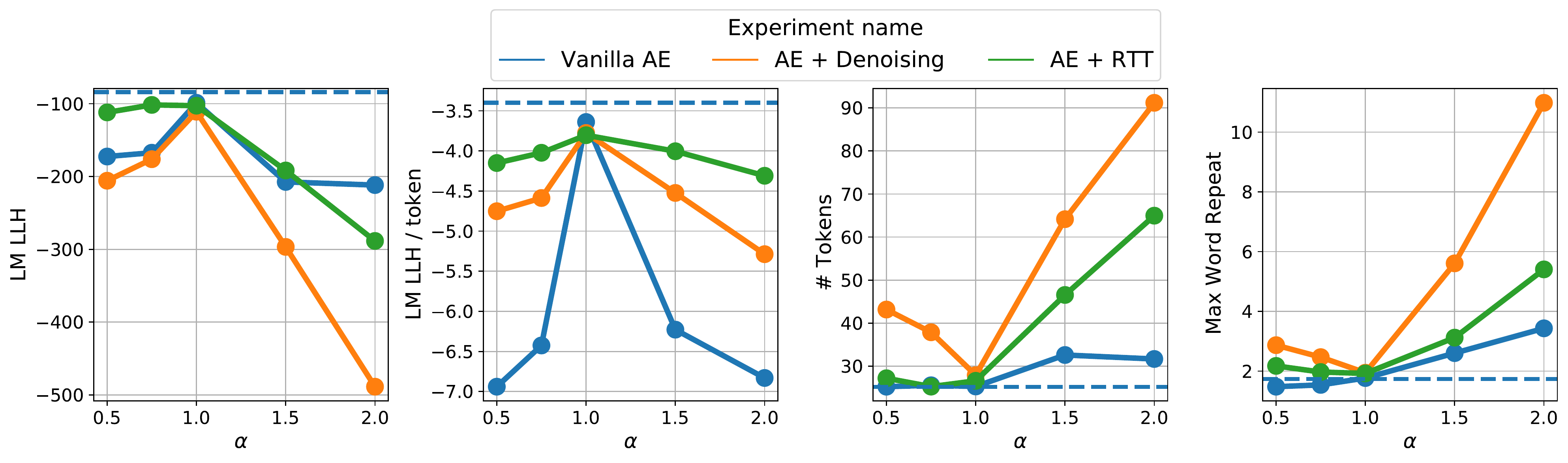}%
}
\vspace{-0.4cm}
\subfloat{%
  \includegraphics[clip,width=\linewidth]{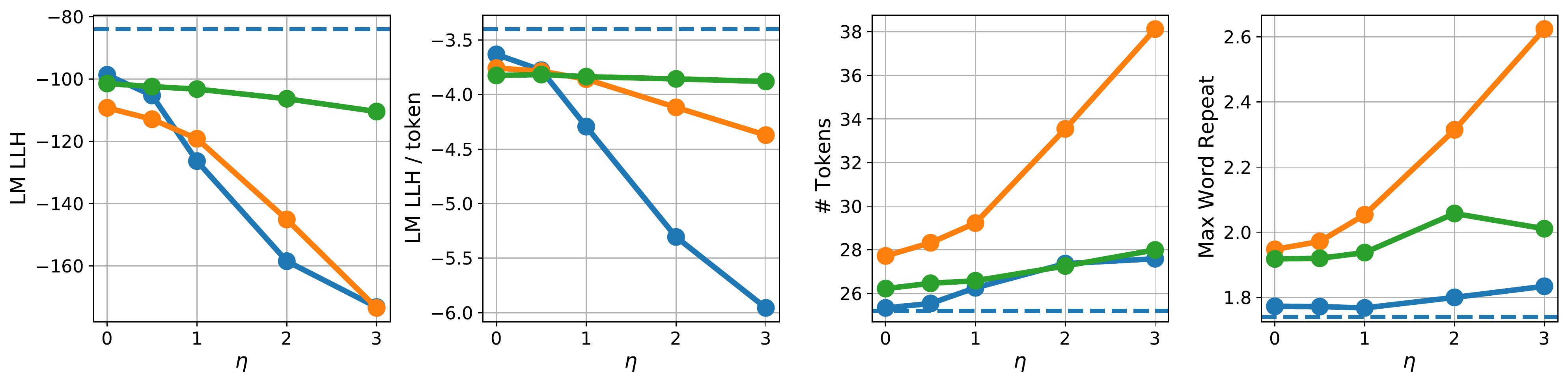}%
}
\caption{Fluency metrics for a fixed bottleneck size of 128 with interpolation (top row) or noise (bottom row). The dotted line corresponds to the statistis computed on 100,000 sentences from C4S where the likelihood is computed with an off-the-shelf T5 model.}
\label{fig:interpolation-noise-alpha-fluency}
\end{figure*}

Figure~\ref{fig:interpolation} shows the results of the different models for an interpolated embedding $z = 0.5 \times z_1 + 0.5 \times z_2$ with $z_1$ and $z_2$ the embeddings of two sentences (sampled from the evaluation dataset of C4S) in function of the bottleneck size while Figure~\ref{fig:interpolation-noise-alpha-fluency} reports the results for a bottleneck size fixed at 128 for different values of interpolation (top row) or noise (bottom row). The full results can be found in Figure~\ref{fig:alpha-interpolation-length} (Appendix~\ref{app:interpolation-quantitative}) for the interpolation and in Figure~\ref{fig:alpha-noise-length} (Appendix~\ref{app:noise-full-experiments}). In all the figures, the dotted line corresponds to the scores computed on 100,000 sentences from the validation dataset of C4S. The likelihood and the likelihood per token are computed with a pre-trained T5 base model (without any bottleneck) that was finetuned on 1M steps on a standard auto-encoding task.

\paragraph{Length and Max Word Repeat} The number of tokens and the number of repeated word for Vanilla AE are close for all bottleneck sizes to the metrics computed on the training distribution. For bottleneck sizes up to 128, RTT also produces sentences that have similar metrics than the sentences in the training distribution. With denoising, the produced sentences are longer and have a larger maximum word repeat than the sentences in the training data distribution. The gap grows when the bottleneck size increases. This indicates that denoising creates sentences more dissimilar with the training distribution than RTT which tends to produce sentences close to the ones in the dataset in term of length. This observation is confirmed in the additional experiments of Section~\ref{sec:denoising} where the Figure~\ref{fig:denoising} shows that increasing the word dropout probability lead to longer sentences. \\ 
We can also note that for bottleneck sizes higher than 128, for both RTT and denoising the mean length of the produced sentences increase significantly. For AE + RTT, the mean length went up from 25 for a bottleneck size 128 to 45 for a bottleneck size 512 while for AE + Denoising the mean length went up from 45 to 60. As a point of comparison, 97\% of the sentences in eval C4S have less than 60 tokens, hence the sentences produced by AE + Denoising are too long to be seen as natural. 
\paragraph{Likelihood} Vanilla AE produces sentences that are short but not very plausible (i.e. medium LM LLH and low LM LLH / token), AE + Denoising produces sentences that are lengthy but are plausible (i.e. low LM LLH but high LM LLH / token), and AE + RTT produces sentences that are both of reasonable size (around the mean length of the dataset) and are plausible (i.e. high LM LLH and high LM LLH / token) especially for bottleneck sizes up to 128.
\looseness=-1
\paragraph{Qualitative examples} These results are confirmed in the qualitative examples in the Section~\ref{app:interpolation-qualitative} where for a bottleneck size equals to 128, the decoded sentences are shown for different values of interpolation. For the Vanilla AE, the decoded sentences from the interpolated embeddings tend to be of reasonable size and are either one of the two anchors sentences or they do not make sense especially when the interpolation weights are at 60\%/40\%. The sentences produced by AE + Denoising are more plausible but they are lengthy especially for the interpolation weights at 60\%/40\%. AE + RTT produces the best sentences in term of interpolation smoothness as the sentences are both plausible and of reasonable size. 
\looseness=-1
\paragraph{Takeaway} AE + RTT (with a bottleneck size up to 128) satisfies the best the \textit{fluency} property as it generate sentences that are both plausible and of similar length to the ones in the data distribution. The other models lack one of these two features: AE + Denoising produces plausible sentences but the denoising perturbation creates sentences too lengthy compared to the data distribution while Vanilla AE produces unlikely sentences which are of similar length with the ones in the data distribution. In addition, the bottleneck size is a crucial parameter for models using RTT or denoising: for bottleneck sizes of 256 and 512, the produced sentences are in average abnormally long compared to the ones in C4S.

\subsection{Semantic Structure}
\label{sec:semantic-structure}

In this section we investigate the semantic structure of the sentence representation spaces. In other words, we check if nearby embeddings produce distribution of sentences with semantic similarities as detailed and formalized in Section~\ref{sec:properties}.

\paragraph{Qualitative interpolation} An additional observation that can be done for the qualitative examples in Section~\ref{app:interpolation-qualitative} is that the sentences produced by AE + RTT, AE + Denoising are semantically close to the anchor sentences. Especially, the models learned with RTT produce a smooth interpolation of the meaning of the anchor sentences. For instance, in Table~\ref{interpolation-ae-rtt}, when the interpolation weights are close to 50\% (second row of the table), AE + RTT produced the sentence \textit{There are kids that are playing with the train.} which captured the \textit{train} word from the first anchor sentence \textit{There are children watching a \textbf{train}.} and also captures the play part from the second anchor sentence \textit{The little girl \textbf{plays} with the toys.}.

\paragraph{Jeffreys divergence} 
\begin{figure}[h]
\includegraphics[width=\columnwidth]{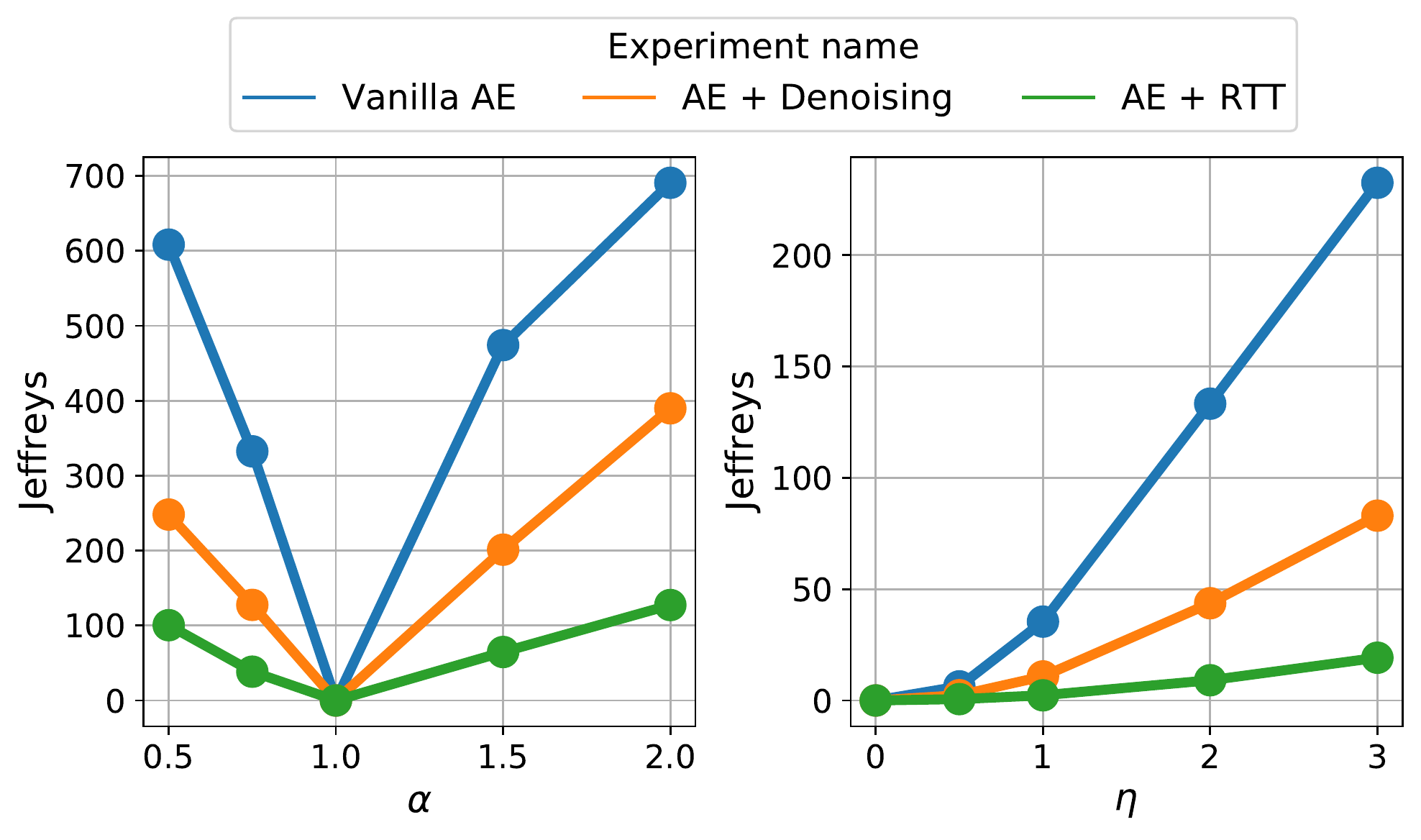}
\caption{Jeffreys divergence between an anchor embedding and an interpolated (right) or noisy (left) embedding. The models were learned with a bottleneck size of 128.} 
\label{fig:jeffreys}
\end{figure}
To quantitatively assess this semantic similarity, we computed the Jeffreys divergence between the distribution of sentences decoded from an anchor embedding and an embededding further and further from the anchor. Figure~\ref{fig:jeffreys} shows that the Jeffreys divergence increases at a lower rate for the models using RTT or denoising while Vanilla AE have a large increase rate of the Jeffreys divergence. In comparison with denoising, RTT models have the lowest increase rate of Jeffreys divergence which indicates that the perturbation coming from RTT bring more semantic structure than the one from denoising.

\paragraph{Topic convex hull} To further understand the semantic structure at the topic level, we uniformly (with a Dirichlet distribution) sample vectors  from a \textit{topic convex hull}. To do so, we first embed few sentences from the same topic as shown in Table~\ref{convex-hull-music-one-col} for the music topic and in Table~\ref{convex-hull-football-one-col} for the football topic and then randomly sample embeddings from the convex hull of the set of the embedded sentences. These sampled embeddings are then decoded to produce the sentences. 

We observe that: Vanilla AE produces sentences that don't make sense and are not on the right topic, AE + Denoising produces sentences on the right topic but part of them don't make sense and are lengthy while AE + RTT produces sentences that both make sense and are on topic. For instance, in Table~\ref{convex-hull-music-one-col}, AE + Denoising produces lenghty sentences that seem to make sense but do not e.g. \textit{As for this song and which songs I love the best music.}. AE + RTT produces shorter sentences that are on topic and differ from the anchor sentences like: \textit{Then you have to have a playback of the music.} or \textit{Name the musical epitaph?}.

\paragraph{Clustering} Given that the models nearly satisfy the universality property, we can check if the structure of the latent sentence representation space to learn about structure of the decoder. We evaluate the semantic structure of this hidden space via clustering. We embedded sentences that are from several different topics and we check if whether these embeddings are clustered by topic. More details about the embedded sentences can be found in Table~\ref{clusters} (Appendix~\ref{app:clustering}). We computed three clustering scores: the Calinski Harabasz score \citep{calinski1974dendrite} where higher is better, the Davies Bouldin score \citep{davies1979cluster} which has a minimum of 0 and where lower is better, and the Silhouette score \citep{rousseeuw1987silhouettes} which is between -1 and 1 and where higher is better. 

Figure~\ref{fig:clustering} shows that both RTT and denoising help with the clustering metrics. The performances of the models generally decrease with the bottleneck size. This is an other example of the accuracy/structure tradeoff where large bottleneck sizes have better reconstruction accuracies but less structure.

\paragraph{Takeaway} Input perturbations like denoising or RTT are key to exhibit some semantic structure. At the sentence level, we saw thanks to the interpolation examples and the Jeffreys divergence that the models with input perturbations produce sentences that can mix the meaning of the anchors sentences. Notably, AE + RTT performed better than AE + Denoising having a lower Jeffreys divergence and more convincing qualitative examples. At the topic level, we saw thanks to the \textit{topic convex hull} evaluation that AE + RTT produced sentences that are both on topic and make sense, AE + Denoising produces sentences about the right topic but a part of them do not make sense, and Vanilla AE is not able to produce sentences about the right topic nor sentences that make sense. The clustering metrics confirmed these observations as they showed that the sentences from the same topic are most scattered in the sentence representation space for Vanilla AE than for the other models.

\subsection{Additional experiments with RTT}
In this section we investigate two variations: (i) the impact of the having diverse rephrasing, (ii) the impact of combining RTT and denoising.

\textbf{Diversity of rephrasing.}
To better comprehend the impact of the diversity of rephrasing on \vtot{} models, we trained another AE + RTT model on a round-trip translated dataset that was created using beam search \citep{tillmann2003word} for decoding instead of nucleus sampling (p=0.9). Figure~\ref{fig:bleu_bs_ns} (Section~\ref{sec:rtt}) shows that beam search created less rephrasing than nucleus sampling (p=0.9) while Figure~\ref{fig:rtt_bs_vs_nucleus} (Section~\ref{sec:rtt}) shows that the model trained with more diversity in the rephrasing lead to a more structure \vtot{} model.

\textbf{RTT combined with denoising.} RTT and denoising being orthogonal dataset modifications, we trained an AE + RTT + Denoising model that combined both perturbations. Figure~\ref{fig:ae_rtt_denoising} shows that the AE + RTT and AE + RTT + Denoising differ only marginally and that the denoising perturbation as the undesirable effect of producing longer sentences.

\section{Other Related Work}
\paragraph{Unsupervised sentence embedding}
Prior work on sentence embedding mainly focused on either model changes \citep{bowman2015generating, montero-etal-2021-sentence, zhao2018adversarially,cifka2018eval,park-lee-2021-finetuning,li-etal-2020-optimus} or dataset modifications \citep{pmlr-v119-shen20c,montero-etal-2021-sentence}.
\citet{bowman2015generating} showed that for a RNN-based language model using a variational auto-encoder (VAE) \citep{kingma2013auto} creates a more structured latent space than using a vanilla auto-encoder. However, a straightforward implementation of the VAE led to similar results than with the vanilla AE. \textit{KL cost annealing}, \textit{word dropout}, and \textit{historyless decoding} were required to see improvements over the baseline. This limitation is confirmed by other works \citep{zhao2018adversarially, pmlr-v119-shen20c,park-lee-2021-finetuning} which found that VAE and Adversarial Auto-encoder (AAE) \citep{makhzani2015adversarial} can not learn meaningful representations out of the box.
\citet{montero-etal-2021-sentence} is another example of a denoising text auto-encoder. In their work, they used RoBERTa \citep{liu2019roberta} -a pre-trained transformers based model- for the encoder and a single transformer layer \citep{vaswani2017attention} for the decoder. \citet{li-etal-2020-optimus} learned a text VAE with a pre-trained BERT \citep{devlin2018bert} encoder and a pre-trained GPT-2 decoder \citep{radford2019language} and showed that using pre-training reduce the KL vanishing issue faced by text VAE.

\paragraph{Round Trip Translation} Round Trip Translation (RTT) was used in automatic paraphrasing \citep{guo2021automatically,zhang2019paws}, quality estimation \citep{moon-etal-2020-revisiting,lample-etal-2018-phrase}, data augmentation for Machine Translation tasks \citep{vaibhav-etal-2019-improving}. To the best of our knowledge, our work is the first to use RTT in the context of text auto-encoder.

\section{Conclusion}
We introduced four desired properties for a \vtot{} model (universality, diversity, fluency, and semantic structure) and showed with thorough quantitative and qualitative results that learning a T5-based universal sentence auto-encoder using round-trip translations produced the best model across the properties.

\bibliography{anthology,custom}

\newpage
\appendix
\onecolumn

\section{Illustration of the different settings: \ttov{}, \ttot{}, \vtot{}}

\begin{figure*}[htb!]
    \centering
    \includegraphics[width=\linewidth]{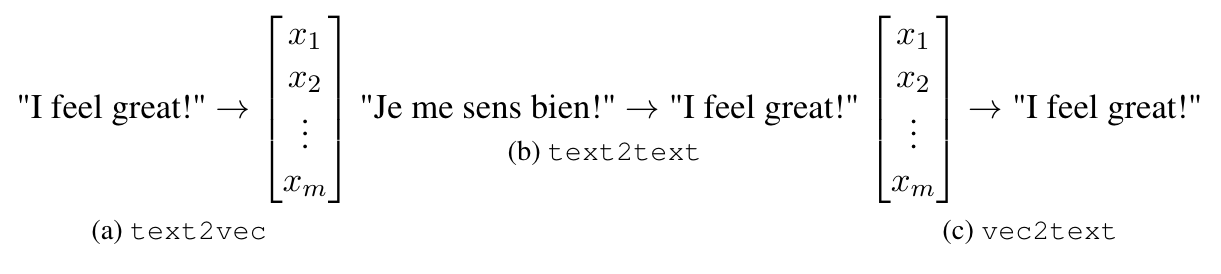}
    \caption{Different settings: This paper focuses on the \vtot{} setting.}
    \label{3figs}
    \label{fig:my_label}
\end{figure*}

\section{Illustration of Round-Trip Translation.}

\begin{table*}[h]
  \centering
  \begin{tabular}{l|l|l}
    \toprule
         & Language     &  \\
    \midrule
    Source  & English &  Please join us for a fun afternoon showing of a feature film. \\
    \hline
    Pivot  & German & \makecell{Nehmen Sie an einem unterhaltsamen Nachmittag teil, \\ bei dem ein Spielfilm gezeigt wird.}       \\  
    \hline
    Round-Trip Translated   & English       &  Take part in a fun afternoon watching a feature film. \\ 
    \bottomrule
  \end{tabular}
  \caption{Illustration of Round-Trip Translation.}
  \label{rtt-table}
\end{table*}

\section{Visualization of the C4S datasets used to train the AE models}

\begin{table*}[h]
  \centering
  \begin{tabular}{lll}
    \toprule
    Dataset     & Input Sentence     & Target Sentence \\
    \midrule
    C4S & I am going to the beach. &   \multirow{5}{*}{I am going to the beach.} \\ \cmidrule(l){1-2} 
    C4S Denoising     & I am \sout{going} to the beach. &        \\ \cmidrule(l){1-2} 
    C4S RTT    & I\textbf{'m on my way} to the beach.       &   \\ 
    \bottomrule
  \end{tabular}
\caption{Visualization of the different datasets. Crossed-out words are dropped by denoising. Bolded words are rephrased by RTT.}
  \label{dataset-table}
\end{table*}

\section{More details on evaluation and training}\label{app:evaluation-details}
\subsection{Evaluation Details}
Evaluating these properties on the whole vector space which is of infinite volume may not be informative as parts of the space have low density of embedded valid English sentences. Instead, we evaluate our models on two bounded set $\mathcal{I}$ and $\mathcal{O}$ where a \ttot{} model should be able to produce valid English sentences. On one hand, the interpolation space $\mathcal{I}$ contains the embeddings of known valid English sentences and the linear interpolation between them: $\mathcal{I} = \{ \alpha E(x_1)+ (1-\alpha) E(x_2) | x_1,x_2 \in \mathcal{E}, \: \alpha \in \{0.5, 0.75, 1, 1.5, 2\}\}$ with $E$ a \texttt{text2vec} model learned jointly with the \vtot{} model $D$. On the other hand, we want to assess if our models have a high density of embeddings that produce valid English sentences with some local semantic coherence around the embeddings of known valid English sentences. Hence, we define a noise space $\mathcal{O} = \{ E(x) + u(x)/||u(x)||_2 \times \sigma \times \eta \: | \: x \in \mathcal{E}, u(x) \sim \mathcal{N}(0, I_{dim(E(x))}), \eta \in \{0, 0.5, 1, 2, 3\}\}$ with $E$ a \texttt{text2vec} model learned jointly with the \vtot{} model, $\sigma$ the standard deviation per dimension of the embeddings, and $\eta$ the noise level. 

\subsection{Approximate Jeffreys divergence}
In our quantitative evaluation of the semantic structure property we compute what we call the approximate Jeffreys divergence at a fix value of interpolation or noise. 

Let's take $\{(s_{1,k},s_{2,k})\}_{k=1,\dots,N}$ a set of valid English sentences that will be used as anchors for the interpolation, their embeddings $\{(e_{1,k},e_{2,k})\}_{k=1,\dots,N}$,  the interpolation between them $\{e_{i, k}=\alpha e_{1,k} + (1-\alpha) e_{2,k}\}_{k=1,\dots,N}$, and a decoded sentence for each interpolation embedding $\{s_{i,k} |s_{i,k} \sim D(e_{i,k})\}_{k=1,\dots,N}$. For $k\in \{1,\dots,N\}$, $J(D(e_{1,k}), D(e_{i,k})) \approx CE(D(e_{i,k}), s_{i,k}) - CE(D(e_{i,k}), s_{1,k}) + CE(D(e_{1,k}), s_{1,k}) - CE(D(e_{1,k}), s_{i,k})$ with $CE(p, t)$ the cross entropy between a distribution $p$ and a target $t$.
\subsection{Training details}
We finetuned both the encoder and the decoder of the different auto-encoders on 300,000 training steps from the publicly available t5.1.1.base checkpoint\footnote{https://github.com/google-research/text-to-text-transfer-transformer/blob/main/released\_checkpoints.md} of a T5 base model ($\sim 250$  million parameters). In this model, the dimension of the embedded tokens is 768. In contrast, the bottleneck sizes that we tried are significantly lower, ranging from 16 to 512. The decoding is done with nucleus sampling \citep{holtzman2019curious} with $p=0.95$ as it provides high quality and diverse sentences.

\section{Additional details on the clustering evaluation and metrics}
\label{app:clustering}
\begin{figure*}[h]
    \centering
    \includegraphics[width=\linewidth]{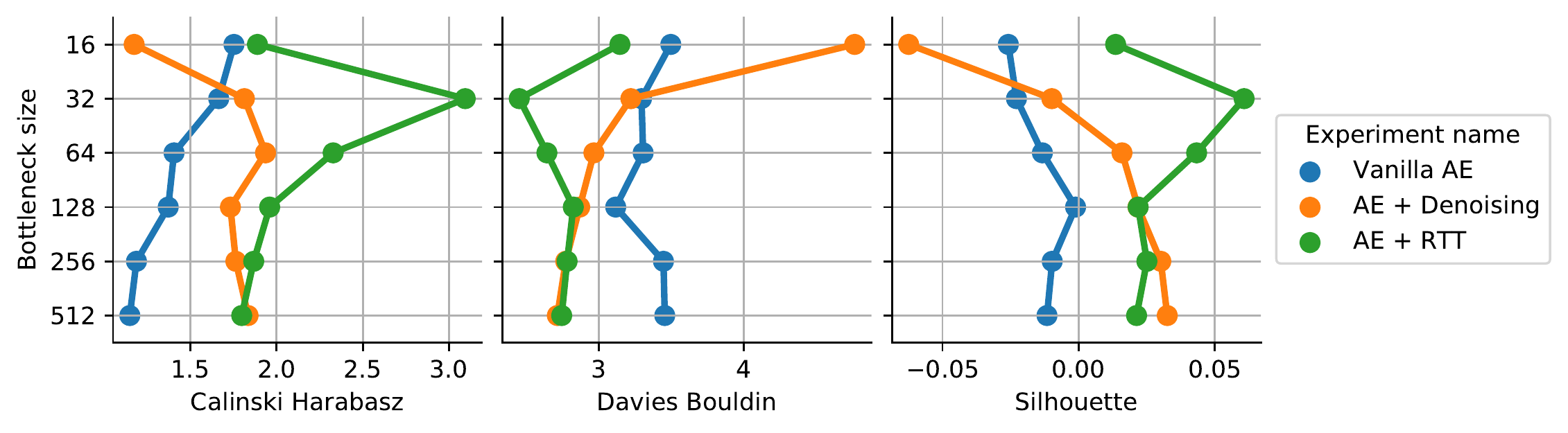}
    \caption{Clustering metrics of embedded topics: the Calinski Harabasz score (where higher is better), the Davies Bouldin score (which has a minimum of 0 and where lower is better), and the Silhouette score (which is between -1 and 1 and where higher is better).}
    \label{fig:clustering}
\end{figure*}

For the clustering evaluation, sentences from different topics were embedded and we computed clustering metrics \citep{davies1979cluster, calinski1974dendrite, rousseeuw1987silhouettes} to see if the sentences from the same topics are closer to each other than sentences from different clusters. The sentences and their associated topics are described in Table~\ref{clusters}. The results are shown in Figure~\ref{fig:clustering}.

\begin{table*}
  \centering
  \caption{Sentences and their topics used for the clustering evaluation.}
  \label{clusters}
  \scalebox{0.95}{
  \begin{tabular}{p{\linewidth}}
    \toprule
    Football Topic \\
    \midrule
    They scored four last week!                      \\
    The goalkeeper made a terrible mistake.          \\
    Have you ever seen this legendary goal?          \\
    I bet you they will win next time.               \\
    Their center-back is not as good as ours.        \\ 
    We have the best striker in all Europe.          \\
    It will be hard to win this year's championship. \\
    \midrule
    Movie Topic \\
    \midrule
    This movie was amazing! \\
    What was this movie about? \\
    When can I go to the cinema? \\
    I don\'t like eating popcorns at the cinema. \\
    The comedy I saw last time was way better. \\ 
    This actress is incredible. \\
    They are a great duo. \\
    \toprule
    Music Topic \\
    \midrule
    I love listening to music. \\
    I played piano for ten years. \\
    What kind of music do you listen to? \\
    I like the rythme and the melody of it. \\
    My favorite artist is from Japan. \\ 
    Could you play a song for us? \\
    Best song ever. \\
    \toprule
    Weather Topic \\
    \midrule
    It rained for two days straight. \\
    Remind me to take my umbrella. \\
    My new coat is perfect against the wind. \\
    It is especially sunny in the morning. \\
    Do you think it is going to get cold at night? \\ 
    Switzerland has the best weather. \\
    I have to check the weather. \\
    \bottomrule
    \end{tabular}}
\end{table*}

\newpage
\section{Comparison between RTT models}
\label{sec:rtt}
\subsection{Diversity of rephrasing}
In this section we compare the structure of \vtot{} models trained on different RTT datasets: the AE + RTT (nucleus p=0.9) which is the model used in the other sections and the AE + RTT (beam search) which is a model learned on a RTT dataset that was created using Beam Search decoding for the translation models.

\begin{figure*}[h]
    \centering
    \includegraphics[width=0.9\linewidth]{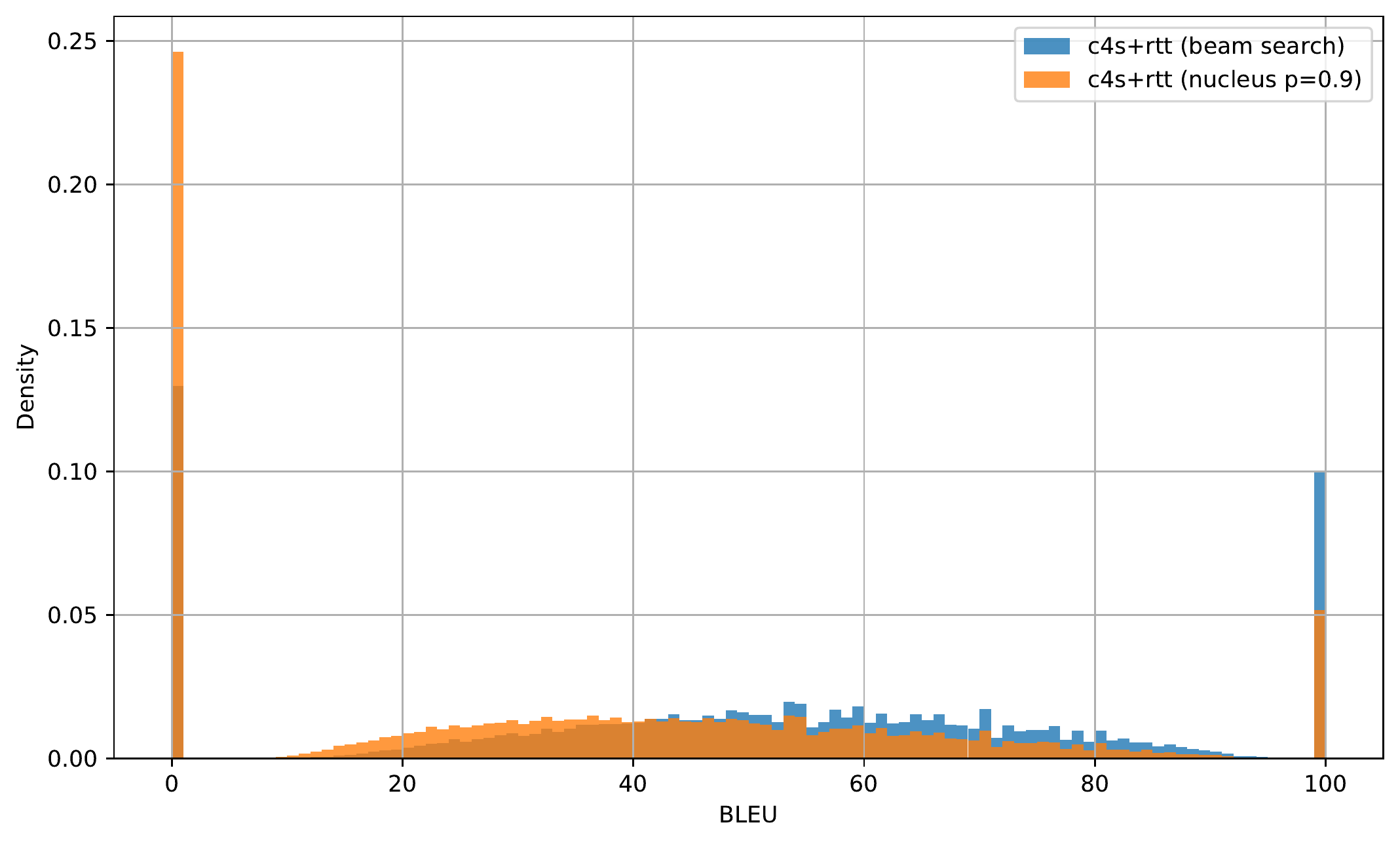}
    \caption{BLEU score between the original sentences and their RTT rephrasing produced with a beam search decoding or a nucleus sampling (p=0.9) decoding.}
    \label{fig:bleu_bs_ns}
\end{figure*}

\paragraph{Datasets} The paraphrasing of the C4S dataset (C4S + RTT) was created with the nucleus sampling (p=0.9) decoding. In addition, we also created the C4S + RTT (beam search) dataset which uses beam search \citep{tillmann2003word} instead of nucleus sampling for the decoding. Figure~\ref{fig:bleu_bs_ns} shows the BLEU score \citep{papineni2002bleu}\footnote{The BLEU score is computed with the open source library sacreBLEU (https://github.com/mjpost/sacrebleu).} between the original sentence and its rephrasing. The rephrasing produced with beam search are less diverse than the ones produced with nucleus sampling (p=0.9): for nucleus sampling, the distribution of the BLEU scores is shifted compared to beam search and with nucleus sampling only 5\% of the rephrasing have a BLEU score of 100 while it is 10\% with beam search. We can note that the high percentage of 0\% BLEU score is an artifact of the fact that we are computing the BLEU score at the sentence level and not for an entire corpus as it was meant to be. Despite that, BLEU at the sentence level is a good indicator to distinguish the amount of paraphrasing.

\paragraph{Results}

\begin{figure*}[h]
    \centering
    \includegraphics[width=\linewidth]{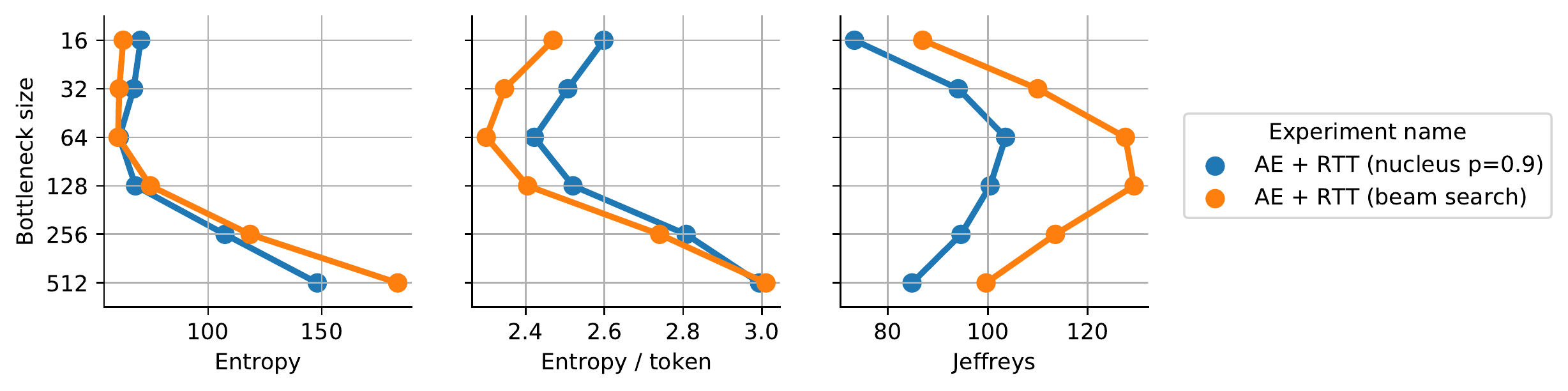}
    \includegraphics[width=\linewidth]{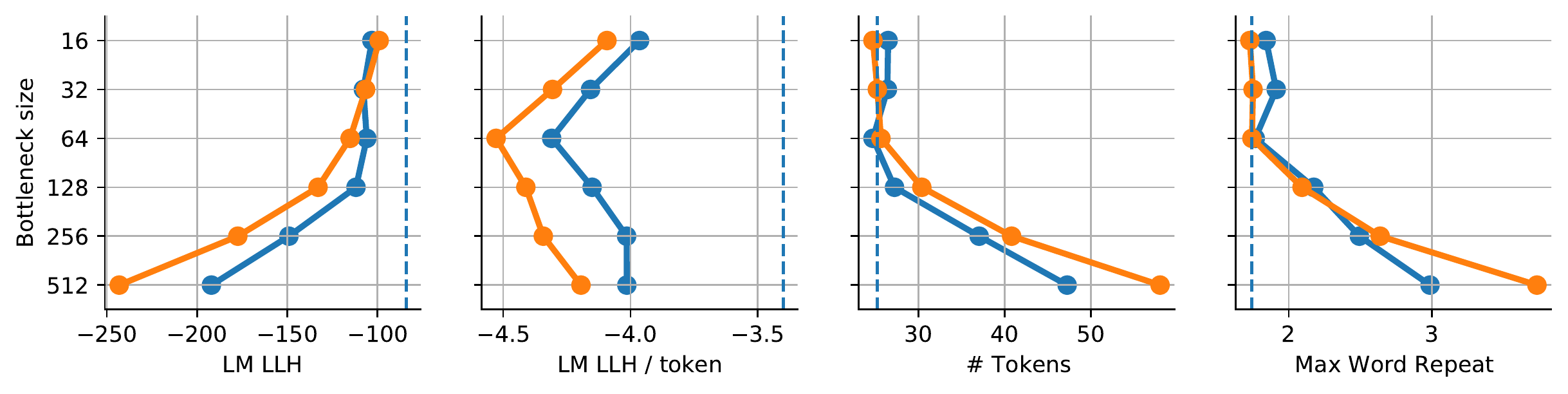}
    \caption{Impact of different RTT datasets on the learned \vtot{} models on interpolated embeddings $z = 0.5 \times z_1 + 0.5 \times z_2$ with $z_1$ and $z_2$ the embedding of two sentences (from the evaluation dataset of C4S). The dotted lines correspond to the statistics computed on 100,000 sentences from C4S.}
    \label{fig:rtt_bs_vs_nucleus}
\end{figure*}

Figure~\ref{fig:rtt_bs_vs_nucleus} shows the comparison between the AE + RTT (nucleus p=0.9) and the AE + RTT (beam search) models for the different properties of sentences decoded from a the mean embedding of two randomly picked sentences. \\
\textbf{Fluency property.} Compared with AE + RTT (beam search), AE + RTT (nucleus p=0.9) produces sentences that have a mean length closer to the ones in C4S especially for the highest values of bottleneck size (256 and 512). AE + RTT (nucleus p=0.9) produces sentences with a higher likelihood and likelihood per token compared to  AE + RTT (beam search). \\
\textbf{Semantic Structure property.} The Jeffreys is lower with  AE + RTT (nucleus p=0.9) than  AE + RTT (beam search) which shows that the sentences produced with AE + RTT (nucleus p=0.9) are semantically closer to the anchors sentences. \\
\textbf{Diversity property.} The entropy per token is slightly higher with AE + RTT (nucleus p=0.9) while the entropy is lower. The lower entropy can be explained as the produced sentences with  AE + RTT (nucleus p=0.9) are shorter than the ones produced with  AE + RTT (beam search).

\paragraph{Takeaway} AE + RTT (nucleus p=0.9) is better across all the properties than AE + RTT (beam search). Hence, having more rephrasing help to learn better \vtot{} models. 

\newpage
\subsection{AE + RTT + Denoising}

\begin{figure}[h]
    \centering
    \includegraphics[width=\linewidth]{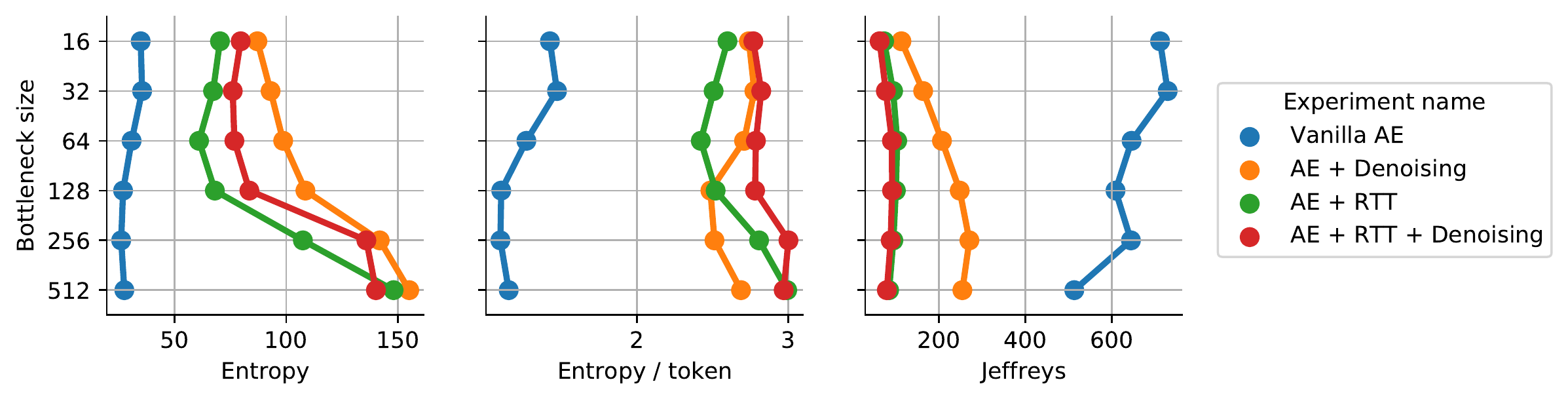}
    \includegraphics[width=\linewidth]{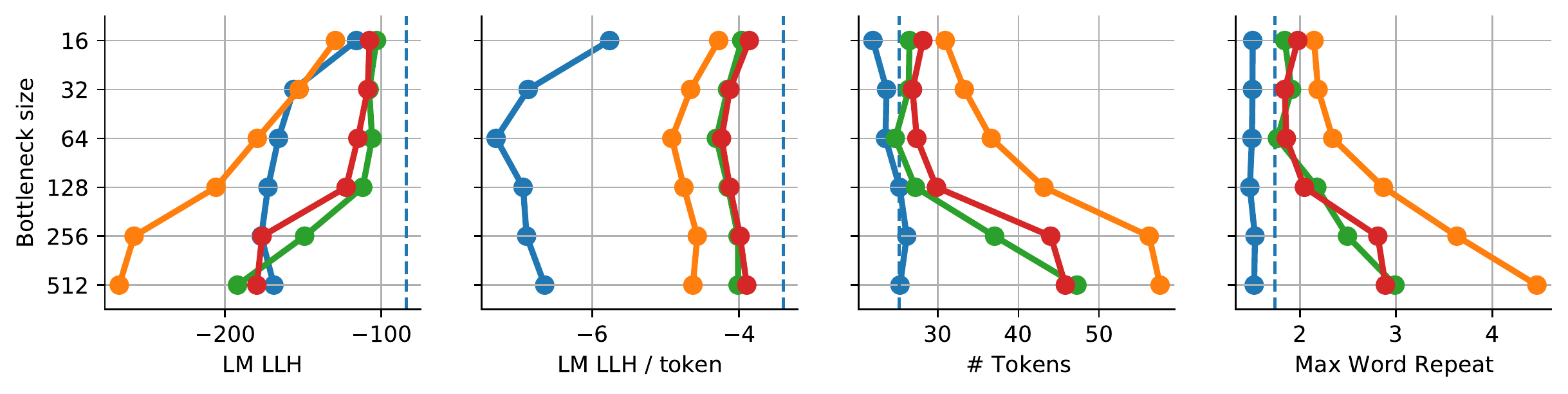}
    \caption{Comparison of the different models on interpolated embeddings $z = 0.5 \times z_1 + 0.5 \times z_2$ with $z_1$ and $z_2$ the embedding of two sentences (from the evaluation dataset of C4S). The dotted lines correspond to the statistics computed on 100,000 sentences from C4S.}
    \label{fig:ae_rtt_denoising}
\end{figure}

RTT and denoising being orthogonal dataset modifications, we trained an AE + RTT + Denoising model that combined both perturbations. As for the AE + Denoising model, the word dropout rate is p=0.2. Figure~\ref{fig:ae_rtt_denoising} shows the comparisons between the different models for an interpolated embedding. AE + RTT and AE + RTT + Denoising have similar performances across all metrics. We can note that AE + RTT + Denoising produces slightly longer sentences than AE + RTT for bottleneck sizes up to 256 which is coherent with our previous observations that denoising has the side effect of creating longer sentences.

\newpage
\section{The ambivalent effect of denoising}
\label{sec:denoising}

\begin{figure*}[h]
    \centering
    \includegraphics[width=\linewidth]{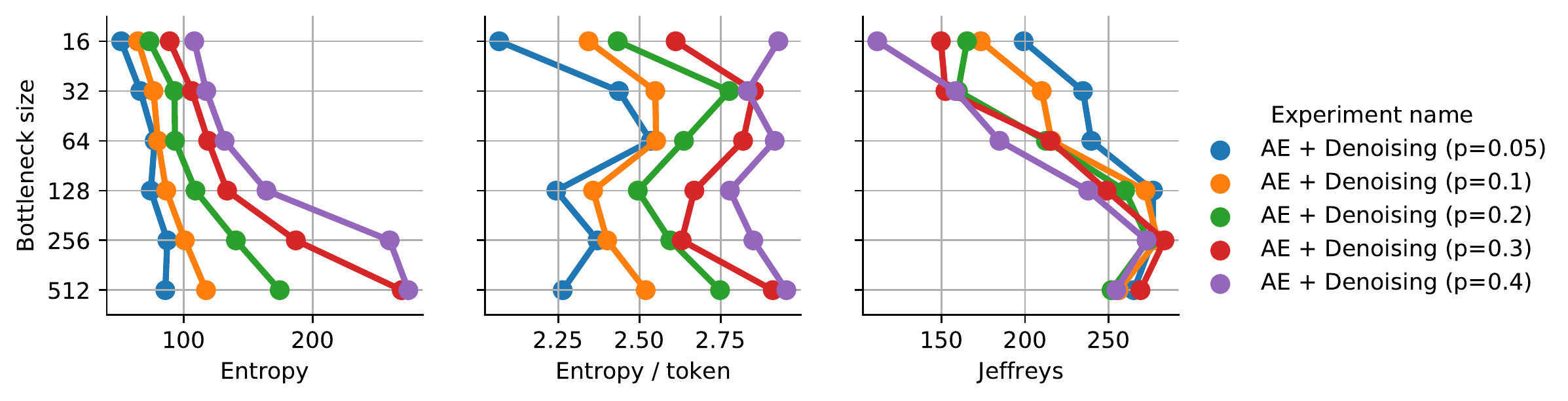}
    \includegraphics[width=\linewidth]{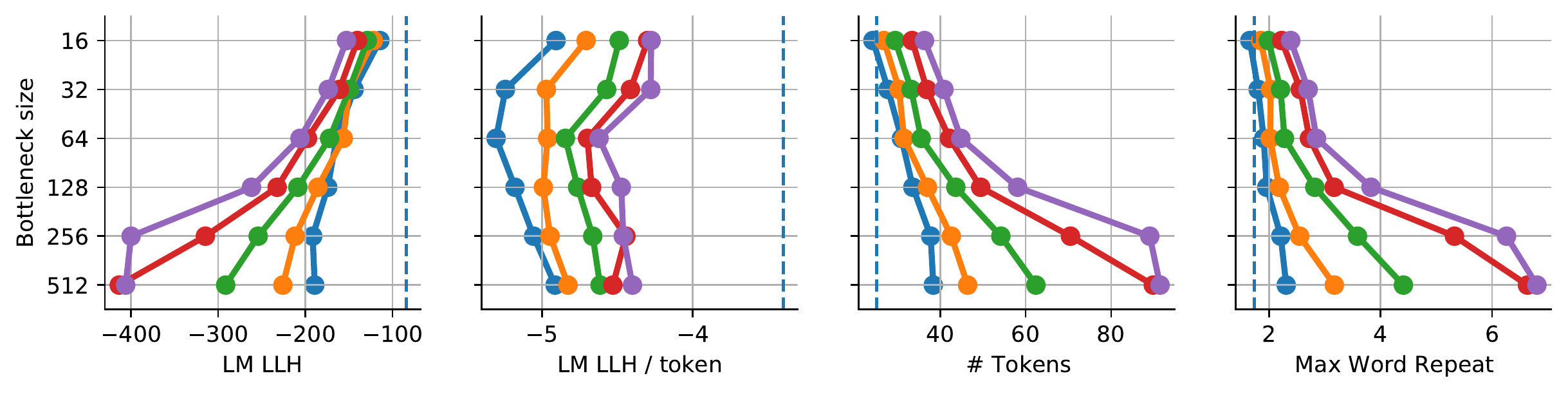}
    \caption{Impact of denoising on \vtot{} models.}
    \label{fig:denoising}
\end{figure*}

To better understand the positive and negative effects of denoising on \vtot{} models, we varied the probability p of word dropout (from 0.05 to 0.4) and evaluated the various denoising models on the previously defined properties. Figure~\ref{fig:denoising} shows the ambivalent effect of denoising. On one hand, a higher word dropout probability rate brings some semantic structure (e.g. a lower Jeffreys) and the produced sentences are more fluent (e.g. a higher likelihood per token). On the other hand, a higher word dropout probability rate also create sentences that are significantly longer (e.g. the mean number of token for a bottleneck size of 128 goes from 35 for p=0.05 to nearly 60 for p=0.4), hence not natural. In comparison, the RTT model produces sentences that are both more fluent and with a mean sentence length closer to the one in the training data distribution.

\newpage
\section{Universality Property Qualitative Examples}
\subsection{AE + RTT model with various bottleneck sizes}
\label{app:universality-bottleneck-size}

\begin{table*}[h]
\scalebox{0.75}{
\begin{tabular}{ll}
\toprule
    Embedded sentence: Our mom painted our old car in red. \\
\midrule
           Our mom already had our old umbrella with mirror in it. & My mother brought our Christmas exteriors in white.  \\
              Our mom got rid of the old dresser from our beehive.  &                         Our mom had painted her old car on black. \\
               Our dad made this favorite bread out of white cake.  &                               Our mother painted our old car red. \\
     You should get flowers from me at Allover Foster's Backpacks.  &                                  Our mom painted our old car red. \\
                 Our mom decorated a vintage campfire for my home.  &                         My family got our hats dyed browns today. \\
       Our mom designed a pine tree for me for that same occasion.  &                               Our mother painted our old car red. \\
            Our mom was setting the chalkboard in our living room.  &              Our mom had painted her old windshield on Christmas. \\
   Our Dad brought home a beautiful orange kernel of bread for us!  &               My mom framed our wintergreens on the front window. \\
                    Our Mother made a pillar from my rustic shelf.  & My family has been painting the hats olive with ours this winter. \\
 bread took our name because of our great-grandparent’s Katherine.  &                               Our mother painted our old car red. \\
\bottomrule
\end{tabular}
}
\caption{Decoding of the embedding of a sentence for AE + RTT with a bottleneck size = 16 (left) and 32 (right).}
\end{table*}

\begin{table*}[h]
\scalebox{0.80}{
\begin{tabular}{ll}
\toprule
    Embedded sentence: Our mom painted our old car in red. \\
\midrule
      Our mum painted our old car red. & Our mom painted our old car in red.\\
 Our mother painted our old car in red.&             Our mom painted our old car red. \\
      Our mom painted our old car red. &             Our mom painted our old car red. \\
   Our mother painted our old car red. &             Our mom painted our old car red. \\
 Our mother repainted our old car red. &             Our mom painted our old car red. \\
    Our Mom painted the old wagon red. & Our mother painted our old car in red paint. \\
      Our Mom painted our old car red. &       Our mother painted our old car in red. \\
      Our mom painted our old car red. &          Our mom painted our old car in red. \\
      Our mom painted our old car red. &          Our mother painted our old car red. \\
   Our mother painted our old car red. &          Our mom painted our old car in red. \\
\bottomrule
\end{tabular}
}
\caption{Decoding of the embedding of a sentence for AE + RTT with a bottleneck size = 64 (left) and 128 (right).}
\end{table*}

\subsection{Bottleneck size = 128 with various models}
\label{app:universality-various-models}

\begin{table*}[h]
\scalebox{0.85}{
\begin{tabular}{ll}
\toprule
 Embedded sentence: Why do you keep arguing that this is the case? & \\
\midrule
Why do you keep arguing that this is the case?  & Why do you keep arguing that this is the case? \\
Why do you keep arguing that this is the case?  & Why do you keep arguing that this is the case? \\
Why do you keep arguing that this is the case?  & Why do you keep arguing that this is the case? \\
Why do you keep arguing that this is the case?  & Why do you keep arguing that this is the case? \\
Why do you keep arguing that this is the case?  & Why do you keep arguing that this is the case? \\
Why do you keep arguing that this is the case?  & Why do you keep arguing that this is the case? \\
Why do you keep arguing that this is the case?  & Why do you keep arguing that this is the case? \\
Why do you keep arguing that this is the case?  & Why do you keep arguing that this is the case? \\
Why do you keep arguing that this is the case?  & Why do you keep arguing that this is the case? \\
Why do you keep arguing that this is the case?  & Why do you keep arguing that this is the case? \\
\bottomrule
\end{tabular}
}
\caption{Decoding of the embedding of a sentence for Vanilla AE (left) and Denoising AE (right) with a bottleneck size = 128.}
\end{table*}

\begin{table*}[h]
\begin{tabular}{l}
\toprule
    Embedded sentence: Why do you keep arguing that this is the case? \\
\midrule
         Why do you keep arguing that this is the way it is? \\
                Why do you repeatedly pretend it does this? \\
                 Why do you keep arguing that this is true? \\
             Why do you keep arguing that this is the case? \\
                   Why do you keep arguing that this is so? \\
 Why do you go on over and over that that this is the case? \\
                  Why do you keep arguing that is the case? \\
                      Why do you argue this is always true? \\
       Why do you keep making claims that this is the case? \\
                    Why are you still arguing against that? \\
\bottomrule
\end{tabular}
\caption{Decoding of the embedding of a sentence for AE + RTT with a bottleneck size = 128.}
\label{decoding-sentence-ae-rtt}
\end{table*}

\newpage
\section{Qualitative interpolation between two sentences}

\label{app:interpolation-qualitative}
In this section, we check if an example of interpolation between two anchor sentences for the various models. We take the weighted average of the embedding of two sentences (Sentence A and Sentence B) and decode (5 times) the embedding $z=\alpha E(A) +(1-\alpha)E(B)$ with $E$ an encoder learned jointly with the \vtot{} models and $\alpha \in \{ 0., 0.2, 0.4, 0.6, 0.8, 1.\}$.

As shown in the Table~\ref{interpolation-vanilla-ae}, for Vanilla AE, traversing the latent space produces sentences that are either the exact anchor sentences (Sentence A or B) or do not make sense (second row of Table~\ref{interpolation-vanilla-ae}).

Table~\ref{interpolation-ae-denoising} shows the sentences produced by AE + Denoising for the interpolation between two sentences. For $\alpha \in \{0, 1\}$, the anchor sentences are either perfectly reconstructed or some small variations can occur (i.e. \textit{several} or \textit{some} are added to the original sentence for $\alpha=1$). For $\alpha \in \{ 0.4, 0.6\}$ (second row of Table~\ref{interpolation-ae-denoising}), i.e. when the interpolated embedding is the further away from the anchor embeddings, the produced sentences are very long as noticed by the quantitative metrics and some of the sentences do not really make sense like \textit{This little girl knows little children and playswith the toys}. 

Table~\ref{interpolation-ae-rtt} shows the sentences produced by AE + RTT for the interpolation between two sentences. For $\alpha \in \{0, 1\}$, the anchor sentences are either perfectly reconstructed or are rephrased e.g. \textit{plays} is changed with \textit{is playing}, \textit{watch} is rephrased with \textit{watching} or \textit{who observe}, and \textit{children} is changed to \textit{kids}. For $\alpha \in \{ 0.4, 0.6\}$ (second row of Table~\ref{interpolation-ae-rtt}), i.e. when the interpolated embedding is the further away from the anchor embeddings, the produced sentences are of reasonable length, make sense, and some mix the two anchor sentences e.g. \textit{There are kids that are playing with the train} combines the \textit{playing} from the sentence A and the \textit{train} from sentence B.

\begin{table*}[h]
  \caption{A qualitative example of interpolation with the Vanilla AE model. \\ Sentence A: The little girl plays with the toys. Sentence B:  There are children watching a train.}
  \label{interpolation-vanilla-ae}
  \centering
  \scalebox{1}{
  \begin{tabular}{p{0.45\linewidth}p{0.45\linewidth}}
    \toprule
    1.0*E(A) + 0.0*E(B) & 0.8*E(A) + 0.2*E(B) \\
    \midrule
    The little girl plays with the toys. &  The little girl plays with the toys. \\
    The little girl plays with the toys. &  The little girl plays with the toys. \\
    The little girl plays with the toys. &  The little girl plays with the toys. \\
    The little girl plays with the toys. &  The little girl plays with the toys. \\
    The little girl plays with the toys. &  The little girl plays with the toys. \\
    \midrule
    0.6*E(A) + 0.4*E(B) & 0.4*E(A) + 0.6*E(B)  \\
    \midrule
        The little girl are observing thehealth. &  There are little children watching a kam force. \\ 
    The little girl are watch the phenomena.  &                  In children are the worse seen. \\
 The little girl are recordings where a the.  &              In children are the mentioned play. \\
                 The little girl are drinks.  &         In children are the least ride to brain. \\
        The little girl are watching agents.  &              There are little children watching. \\
    \midrule
    0.2*E(A) + 0.8*E(B) & 0.0*E(A) + 1.0*E(B)  \\
    \midrule
 There are children watching a train. &  There are children watching a train. \\
 There are children watching a train. &  There are children watching a train. \\
 There are children watching a train. &  There are children watching a train. \\
 There are children watching a train. &  There are children watching a train. \\
 There are children watching a train. &  There are children watching a train. \\
        
    \midrule
  \end{tabular}
  }
\end{table*}

\begin{table*}[h]
  \caption{A qualitative example of interpolation with the AE + Denoising model. \\ Sentence A: The little girl plays with the toys. Sentence B:  There are children watching a train.}
  \label{interpolation-ae-denoising}
  \centering
  \scalebox{1}{
  \begin{tabular}{p{0.45\linewidth}p{0.45\linewidth}}
    \toprule
    1.0*E(A) + 0.0*E(B) & 0.8*E(A) + 0.2*E(B) \\
    \midrule
    The little girl plays with the toys. &            The little girl plays with the toys. \\
    The little girl plays with the toys. &   The little girl plays with the kids and toys. \\
    The little girl plays with the toys. &            The little girl plays with the toys. \\
    The little girl plays with the toys. &            The little girl plays with the toys. \\
    The little girl plays with the toys. &            The little girl plays with the toys. \\
    \midrule
    0.6*E(A) + 0.4*E(B) & 0.4*E(A) + 0.6*E(B)  \\
    \midrule
 This is where the little children get to play and dress up while they enjoy playing with the funky team play blocker pancakes and swimming in a pool. &  This year the children are finding the same magic with that idea of seeing what the reddish light was singing on the weather when you were on a bike.\\ 
 This little puppeteer is cat food for kids and the children while playing. & We know that your little boy and toddler children are always watching the odd grasshopper and the hippopotamus. \\
 This little girl has tiny little children playing the jigsaw. & It sounds like your children are watching a little snow girl playing with the lights. \\
 This week the girl and her little cubby children played with the wall swings and the basket spins and spins around with the chandeliers. &       It means that the children are always looking for a good song while watching the TV or the sleeping girl. \\
 This little girl knows little children and plays with the toys. &  And some children are there to let out a lively singing and playing every day with the walking and the flying shows. \\
    \midrule
    0.2*E(A) + 0.8*E(B) & 0.0*E(A) + 1.0*E(B)  \\
    \midrule
    That way children are watching a piano play with a train. & There are some children watching a train. \\
  That day children are watching a softball game on the christmas tree on the Christmas morning train. & There are children watching a train. \\
  That is why children are watching the train on the overtaking train. & There are children watching a train. \\
  That they the children are watching from a snowmobile on a tractor cable car. & There are several children watching a train. \\
  That means children are watching a train with their little dog. & There are children watching a train. \\
        
    \midrule
  \end{tabular}
  }
\end{table*}

\begin{table*}[h]
  \caption{A qualitative example of interpolation with the AE + RTT model. \\ Sentence A: The little girl plays with the toys. Sentence B:  There are children watching a train.}
  \label{interpolation-ae-rtt}
  \centering
  \scalebox{1}{
  \begin{tabular}{p{0.45\linewidth}p{0.45\linewidth}}
    \toprule
    1.0*E(A) + 0.0*E(B) & 0.8*E(A) + 0.2*E(B) \\
    \midrule
     Oh, the little girl playing with the toys. &             This little girl played with the toys. \\
      The little girl is playing with the toys. &              The little child plays with the toys. \\
            The little girl play with the toys. &        The kids use the model to play around with. \\
      The little girl is playing with the toys. &  The little one loves playing with the characters. \\
        The baby girl is playing with the toys. &          The little girl is playing with the toys. \\
    \midrule
    0.6*E(A) + 0.4*E(B) & 0.4*E(A) + 0.6*E(B)  \\
    \midrule
                    Aladdin Watching Children Playing. &                                     Kids rewatch. \\ 
                 The kids are playing with the lights. &               There are children watching the train. \\
 The little one looks at play on the terms of the two. &            The children are playing with the SCENES. \\
                    children play games with the tree. &                          Children watching the race. \\
           Kids are playing and playing with the toys. &      There are kids that are playing with the train. \\
    \midrule
    0.2*E(A) + 0.8*E(B) & 0.0*E(A) + 1.0*E(B)  \\
    \midrule
Those kids are looking at the tops and watching the trains. &      There are children watching a train. \\
There’s a kid watching a train. &  There are children who observe a train. \\
There are children watching a train. &    There are children who watch a train. \\
There are children watching the train. &     There are children watching a train. \\
Having children watching a train. &         There are kids watching a train. \\
        
    \midrule
  \end{tabular}
  }
\end{table*}

\begin{table*}[h]
  \caption{A qualitative example of interpolation with the Vanilla AE model. \\ Sentence A: We can eat whenever you want. Sentence B: Let's not eat yet.}
  \label{interpolation-2-vanilla-ae}
  \centering
  \scalebox{1}{
  \begin{tabular}{p{0.45\linewidth}p{0.45\linewidth}}
    \toprule
    1.0*E(A) + 0.0*E(B) & 0.8*E(A) + 0.2*E(B) \\
    \midrule
     We can eat whenever you want. &   We can eat whenever you want. \\
     We can eat whenever you want. &   We can eat whenever you want. \\
     We can eat whenever you want. &   We can eat whenever you want. \\
     We can eat whenever you want. &   We can eat whenever you want. \\
     We can eat whenever you want. &   We can eat whenever you want. \\
    \midrule
    0.6*E(A) + 0.4*E(B) & 0.4*E(A) + 0.6*E(B)  \\
    \midrule
                    We can eat’t. &            Our can eat’t not. \\ 
 We can’t eat vorbei by everyone. &    Our can’t eat here is Don. \\
      We can’ eat here not wants. &                Our can’t eat. \\
    We can’ eat here not attract. &  Our can’t eat vorbei by you. \\
            We can’t eat those by &                Our can’t eat. \\
    \midrule
    0.2*E(A) + 0.8*E(B) & 0.0*E(A) + 1.0*E(B)  \\
    \midrule
  Let’s not eat yet. &  Let’s not eat yet. \\
  Let’s not eat yet. &  Let’s not eat yet. \\
  Let’s not eat yet. &  Let’s not eat yet. \\
  Let’s not eat yet. &  Let’s not eat yet. \\
  Let’s not eat yet. &  Let’s not eat yet. \\
        
    \midrule
  \end{tabular}
  }
\end{table*}

\begin{table*}[h]
  \caption{A qualitative example of interpolation with the AE + Denoising model. \\ Sentence A: We can eat whenever you want. Sentence B: Let's not eat yet.}
  \label{interpolation-2-ae-denoising}
  \centering
  \scalebox{1}{
  \begin{tabular}{p{0.45\linewidth}p{0.45\linewidth}}
    \toprule
    1.0*E(A) + 0.0*E(B) & 0.8*E(A) + 0.2*E(B) \\
    \midrule
                  We can eat them whenever you want. &  We can eat there whenever you want. \\
                       We can eat whenever you want. &        We can eat whenever you want. \\
     We can eat whatever you like whenever you want. &  We can eat lunch whenever you want. \\
      We can eat whatever time or whenever you want. &        We can eat whenever you want. \\
             We can eat and drink whenever you want. &     We can eat it whenever you want. \\
    \midrule
    0.6*E(A) + 0.4*E(B) & 0.4*E(A) + 0.6*E(B)  \\
    \midrule
We can eat it for breakfast whenever you want. &  She told him that he couldn’t not eat immediately for breakfast yet. \\ 
We can eat together whenever we want. & She suggested we wouldn’t eat for food today either. \\
We can eat pizzas if we please them and wish we had a delightful holiday perhaps. & Let’s eat whenever we can without filling so much with food. \\
We can eat it anytime and whenever you want. & Let’s eat lunch whenever not yet. \\
We can eat here for dinner whenever you want. & She CAN’T EAT when they sit here just yet. \\
    \midrule
    0.2*E(A) + 0.8*E(B) & 0.0*E(A) + 1.0*E(B)  \\
    \midrule
                     Let’s not eat yet. &           Let’s not eat fish yet. \\
                     Let’s not eat yet. &       Let’s not eat heartily yet. \\
            Let’s not eat together yet. &                Let’s not eat yet. \\
 Let’s eat our way through and not yet. &  Let’s not eat all the talks yet. \\
                     Let’s not eat yet. &        Let’s not eat forever yet. \\
        
    \midrule
  \end{tabular}
  }
\end{table*}

\begin{table*}[h]
  \caption{A qualitative example of interpolation with the AE + RTT model. \\ Sentence A: We can eat whenever you want. Sentence B: Let's not eat yet.}
  \label{interpolation-2-ae-rtt}
  \centering
  \scalebox{1}{
  \begin{tabular}{p{0.45\linewidth}p{0.45\linewidth}}
    \toprule
    1.0*E(A) + 0.0*E(B) & 0.8*E(A) + 0.2*E(B) \\
    \midrule
 We can eat whenever you want. &  We can eat anytime you are in need. \\
 We can eat whenever you want. &               We’ll eat if you want. \\
      We can eat, if you want. &             I can eat what you want. \\
 We can eat any time you want. &         We can eat whatever we want. \\
     We eat whenever you want. &           Can eat whenever you want. \\
    \midrule
    0.6*E(A) + 0.4*E(B) & 0.4*E(A) + 0.6*E(B)  \\
    \midrule
 We can eat out whenever you please. &                Lets eat it for free. \\
        We can eat whatever we want. &              Let’s eat as we please. \\
      We can eat however we want to. &            Let’s eat it if you like. \\
     We can eat as long as you like. &                        Let’s eat it. \\
         We can eat it when we want. &  Let's eat what we want a chance of. \\
    \midrule
    0.2*E(A) + 0.8*E(B) & 0.0*E(A) + 1.0*E(B)  \\
    \midrule
 Let’s not eat them yet. &                Let’s not eat yet. \\
      Let’s not eat yet. &  Let’s eat more, not eating less. \\
     Let's eat it ain't. &                Let’s not eat yet. \\
      Let’s not eat yet. &              Let’s don’t eat yet. \\
      Let's not eat yet. &                Let’s not eat yet. \\
        
    \midrule
  \end{tabular}
  }
\end{table*}

\section{Quantitative interpolation between two sentences}
\label{app:interpolation-quantitative}
Figure~\ref{fig:alpha-interpolation-length} shows the full quantitative metrics for the interpolation between two sentences evaluation. 
\begin{figure}[htb!]
\centering
\caption{Quantitative results on the compact set $\mathcal{I}$ that contains different values of interpolation.}
\label{fig:alpha-interpolation-length}
\subfloat[]{%
  \includegraphics[clip,width=0.85\linewidth]{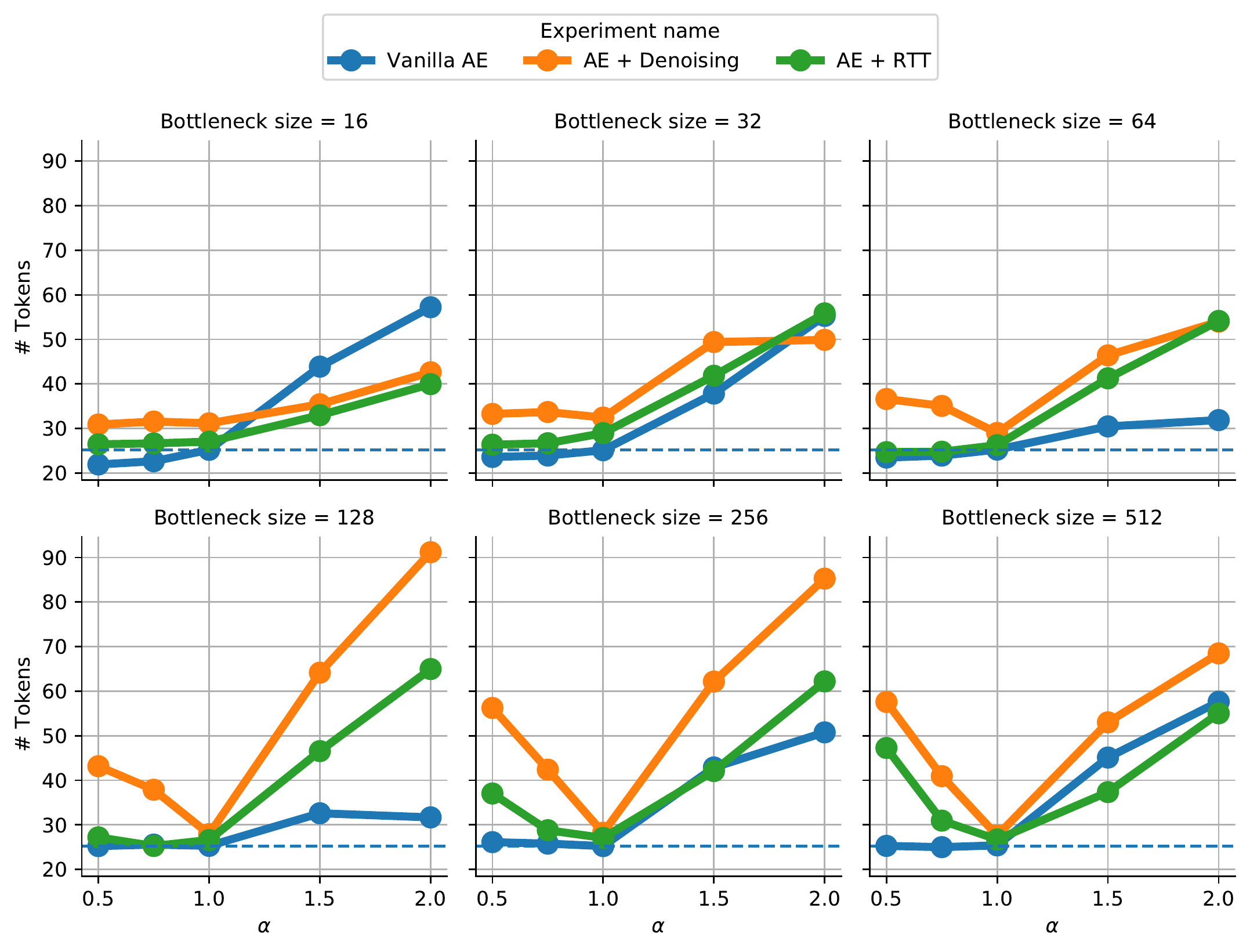}%
}
\vspace{-0.7cm}
\subfloat[]{%
  \includegraphics[clip,width=0.85\linewidth]{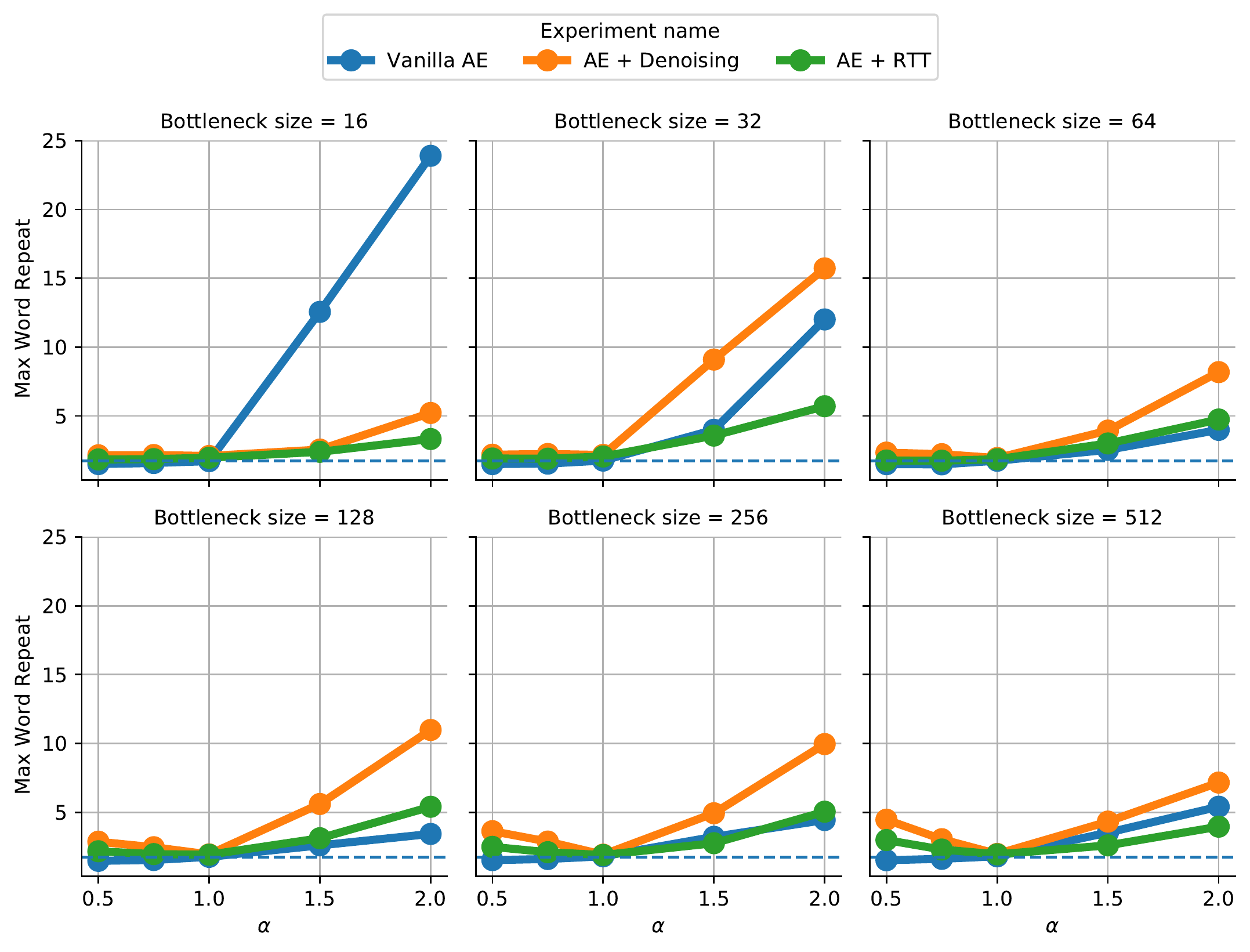}%
}
\end{figure}

\begin{figure}[h]\ContinuedFloat
\centering
\subfloat[]{%
  \includegraphics[clip,width=0.9\linewidth]{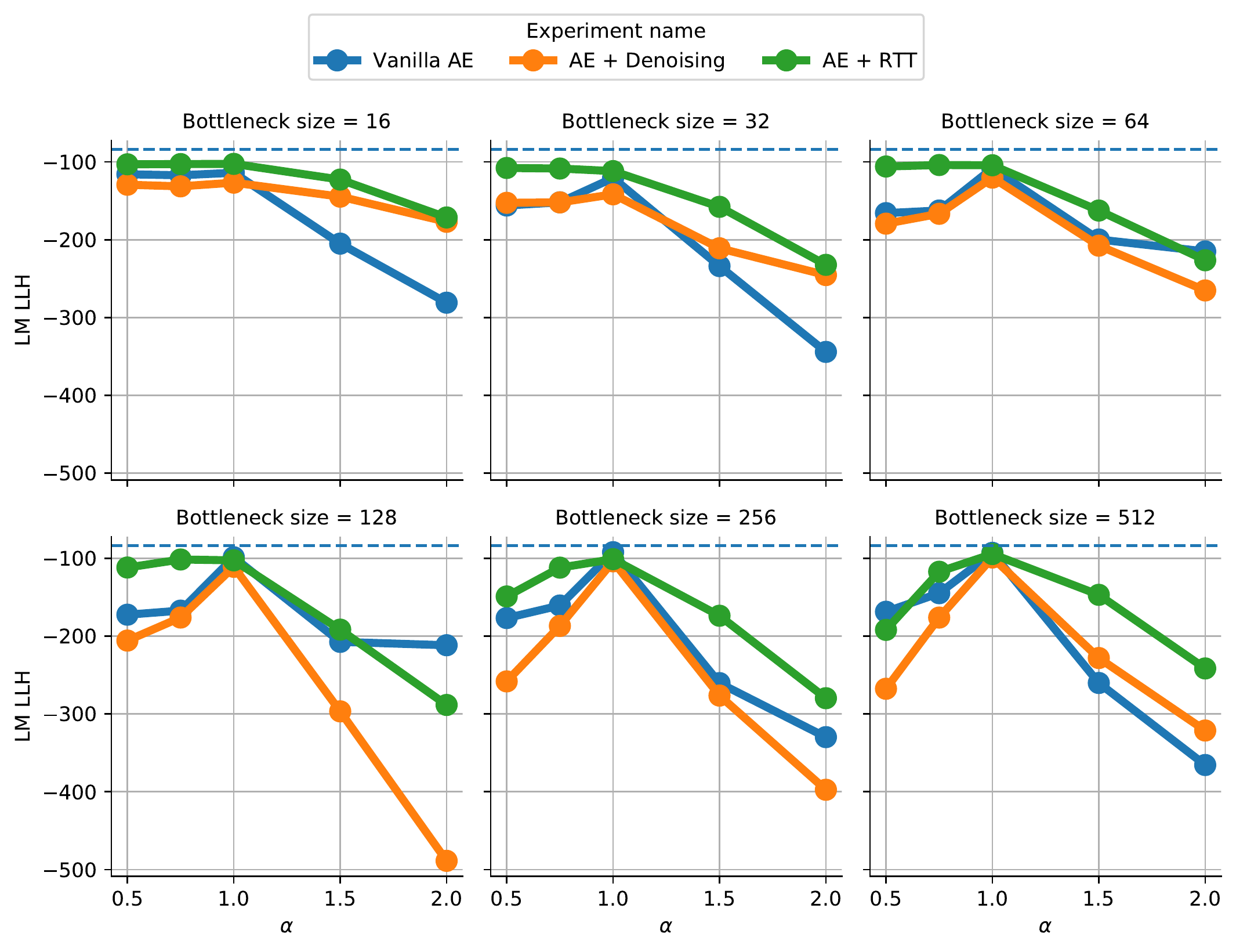}%
}

\subfloat[]{%
  \includegraphics[clip,width=0.9\linewidth]{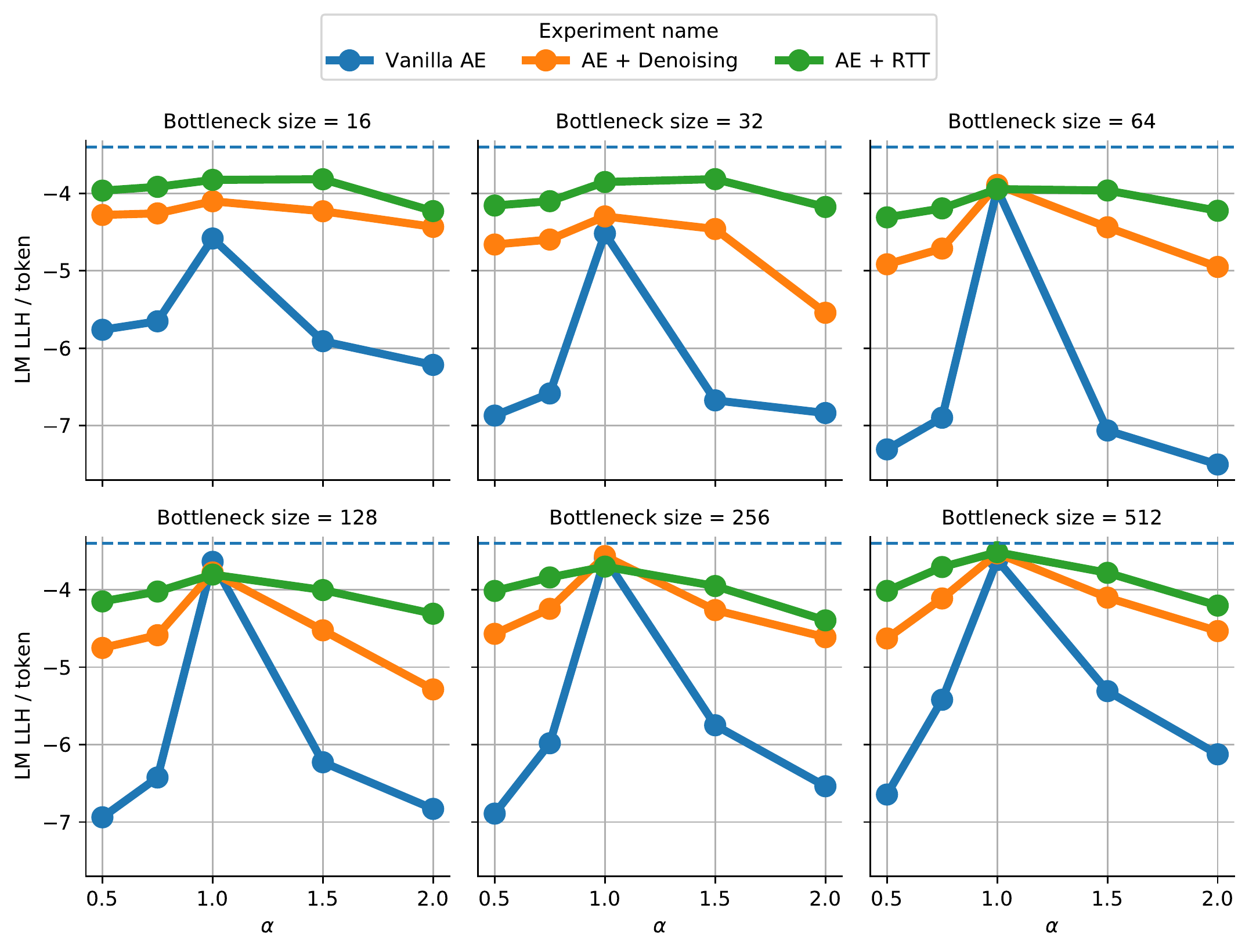}%
}
\label{fig:alpha-interpolation-lm}
\phantomcaption
\end{figure}

\begin{figure}[h]\ContinuedFloat
\centering
\subfloat[]{%
  \includegraphics[clip,width=0.9\linewidth]{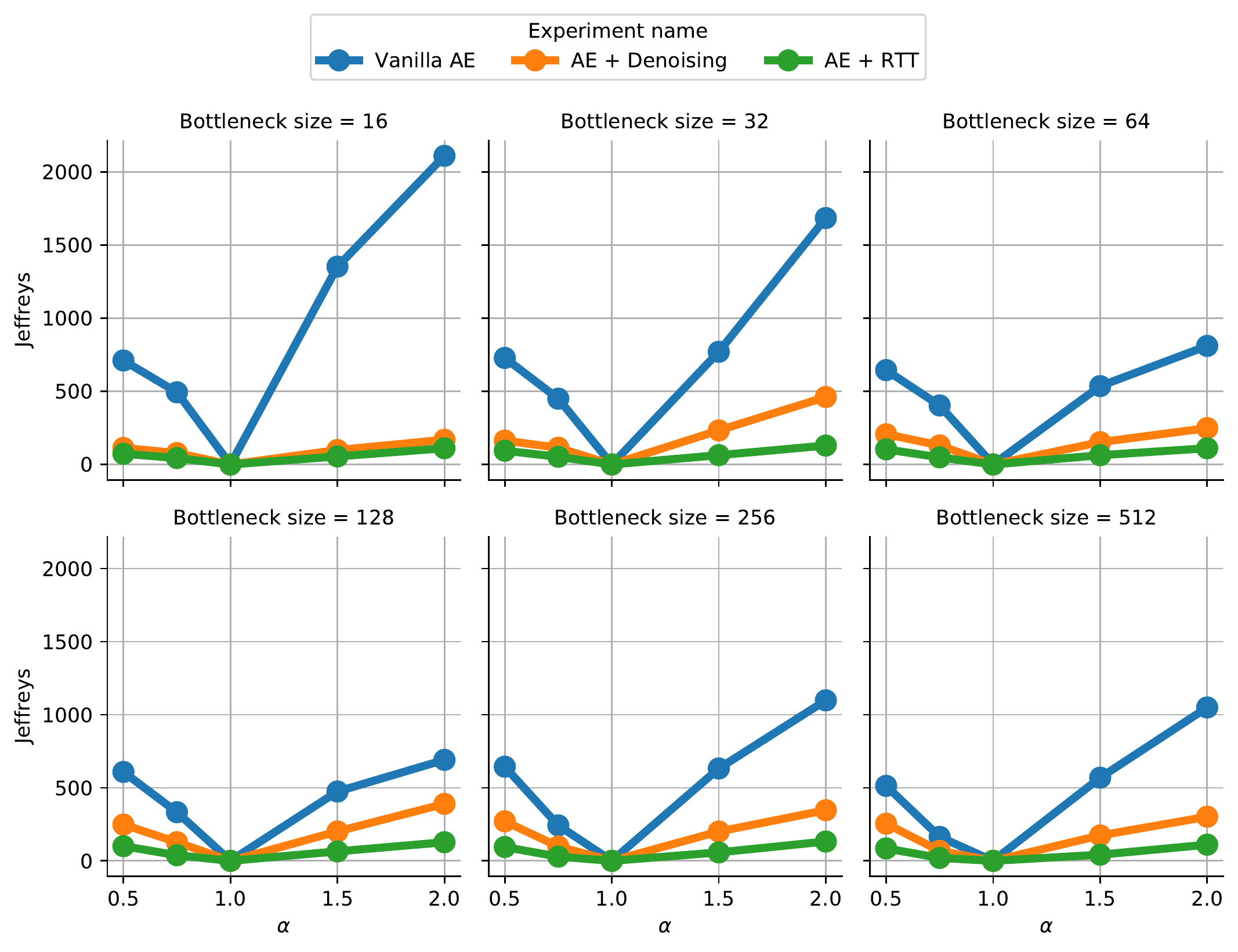}%
}
\label{fig:alpha-interpolation-jeffreys}
\phantomcaption
\end{figure}

\begin{figure}[h]\ContinuedFloat
\centering
\subfloat[]{%
  \includegraphics[clip,width=0.9\linewidth]{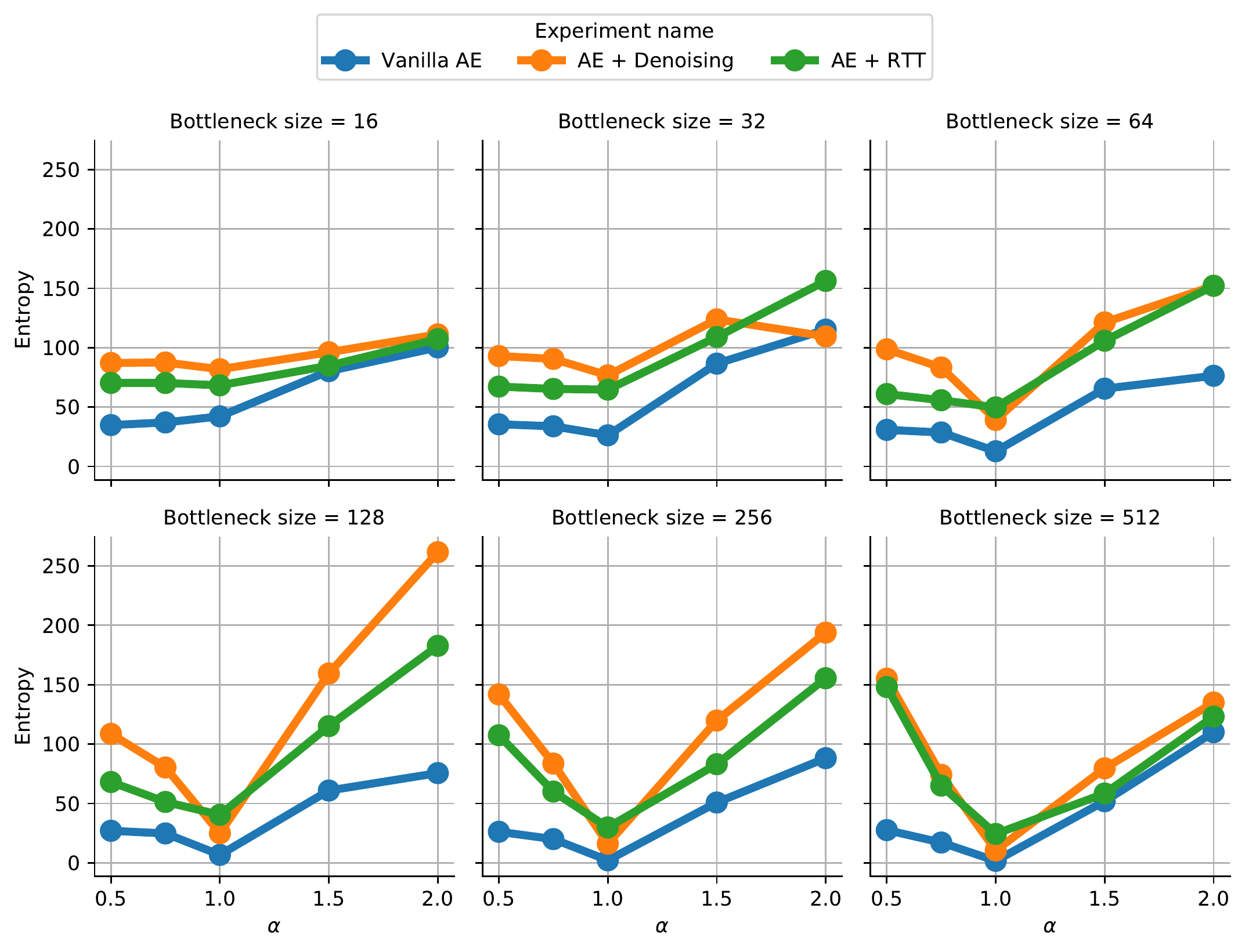}%
}

\subfloat[]{%
  \includegraphics[clip,width=0.9\linewidth]{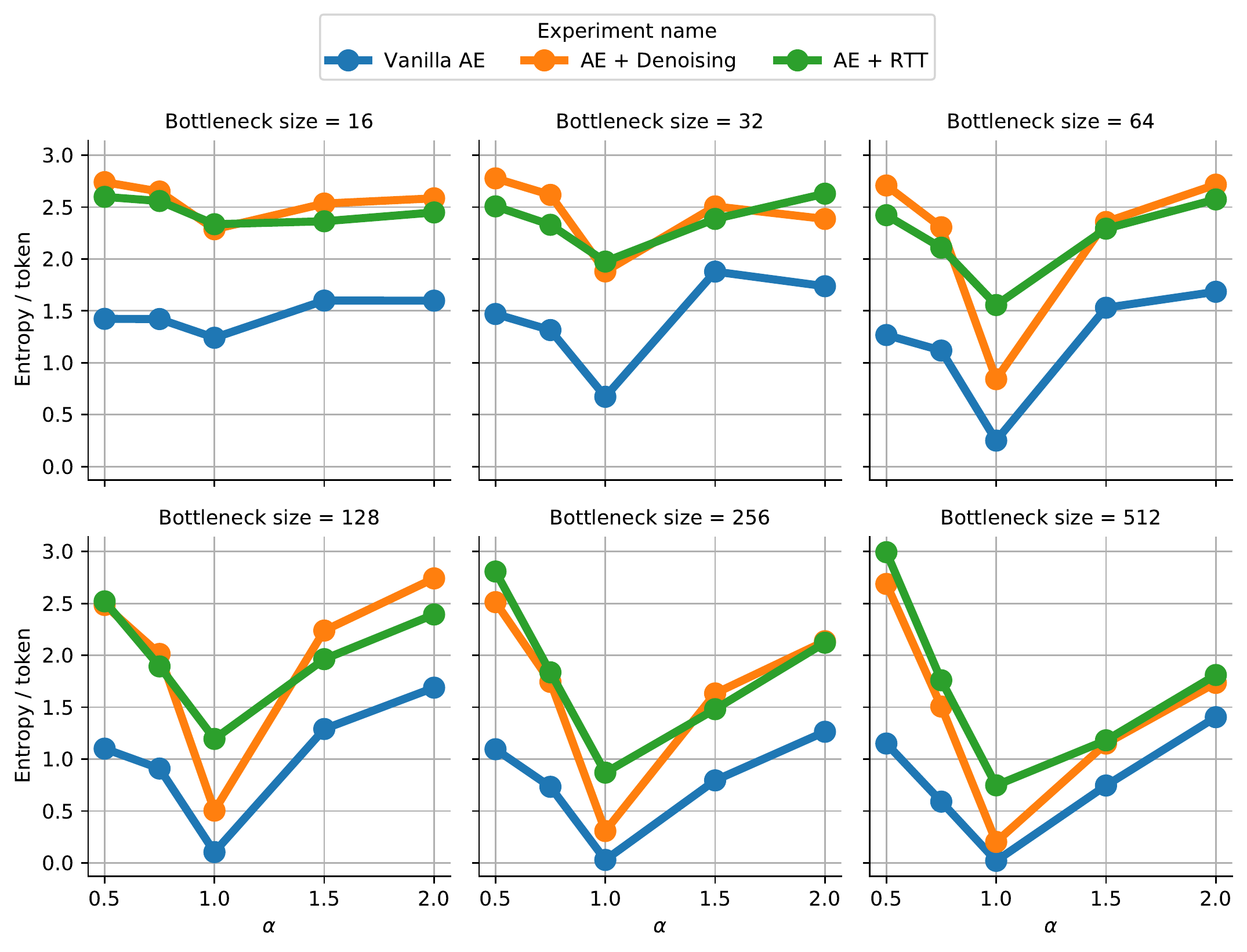}%
}

\label{fig:alpha-interpolation-ent}
\phantomcaption
\end{figure}

\newpage
\section{Semantic Structure Additional Qualitative Examples}
\label{app:semantic-structure}
In this section, we show additional topic convex hull evaluation examples: one for the music topic (Table~\ref{convex-hull-music-one-col}) and one for the football topic (Table~\ref{convex-hull-football-one-col}).

\begin{table*}[h]
  \centering
  \caption{Decoding of vectors sampled randomly within a topic convex hull. The sentences chose in this evaluation are on the music topic.}
  \label{convex-hull-music-one-col}
  \scalebox{0.83}{
  \begin{tabular}{p{\linewidth}}
    \toprule
    Music Topic \\
    \midrule
    I love listening to music. \\
    You played piano for ten years. \\
    What kind of music do you listen to? \\
    We like the rhythm and the melody of it. \\
    My favorite artist is from Japan. \\
    Could you play a song for us? \\
    Best song ever. \\
    \midrule
    Vanilla AE \\
    \midrule
                      She <extra\_id\_26> \\
                           He like Susan. \\
  My use thé than you research’ festival. \\
  My love youtude for the experiments in. \\
    She ability of 30 played at the year. \\
                             My love you. \\
            She popular abilities for me. \\
 She possible://80 lectures to you member \\
   She best applications for the talking. \\
  Our sound like you can eat ministries). \\
  \midrule
  AE + Denoising \\ 
  \midrule
  Everything I do has my kids love playing the music. \\
  As a Kubaian ifa play is a superb song and you want to play it with me? \\
  As a piano player I have played a piano for years. \\
  Have a gift for the kind of music you play? \\
  As the tracker you are playing is a great way to get your music playing right on your big screen. \\
  As for this song and which songs I love the best music. \\
  I love listening to and experiencing music played. \\
                 Would you play a song for me? \\
  What kind of music did you love the best track? \\
  One might think this family member is the perfect Happy times music favorite for the new frogborns of the Californian saxophonists of western. \\
   \midrule
   AE + RTT \\ 
   \midrule
                                               was it an appropriate song? \\
                                               I like the cut of the song. \\
                                     What kind of music do you guys enjoy? \\
                            Then you have to have a playback of the music.\\
                                                    Branded Best of State. \\
                               In my local area I like to listen to music. \\
                                       The one I love most is the podcast.\\
 How lovingly listen to this stuff that I usually do when I want to dance. \\
                                                 Name the musical epitaph?\\
                       A favorite thing is to listen to it, so pretty too. \\
    \bottomrule
  \end{tabular}
}
\end{table*}

\begin{table}[h]
  \centering
  \caption{Decoding of vectors sampled randomly within a topic convex hull. The sentences chose in this evaluation are on the football topic.}
  \label{convex-hull-football-one-col}
  \scalebox{0.85}{
  \begin{tabular}{p{\linewidth}}
    \toprule
    Football Topic \\
    \midrule
    They scored four last week! \\
    The goalkeeper made a terrible mistake. \\
    Have you ever seen this legendary goal? \\
    I bet you they will win next time. \\
    Their center-back is not as good as ours. \\ 
    We have the best striker in all Europe. \\
    It will be hard to win this year's championship. \\
    \midrule
    Vanilla AE \\
    \midrule
                            We have the bestim being vana more. \\
                          And have youm successful to the gold. \\
                                      We have the best striker. \\
               Inhe willVAL 2, your been to support Florida and \\
            We be1\% department’s last you can change the stage. \\
               The goalen held a terrible dividend as richtige. \\
                          And spent their belty will not great. \\
            This beers senior d like the last game excellence," \\
                                 You havet much Freiburg://www. \\
 This have goal youal laughed from the last series," confirmed. \\
  \midrule
  AE + Denoising \\ 
  \midrule
this time you have to bet you have been the best of the cut team this year. \\
We bet you will be the only one who has scored multiple times in the last year! \\
We scored deserve to have proven to be the last rookie in the Westbrook United goal. \\
He hopes the sprinters look like center or aren't as good after all except their matches.” \\
In fact I think Sweden have just a made the last right and probably the worst goal. \\
Our lucky win will make a point and be a great goalkeeper because John has a real talent at the same time. \\
Have you ever seen him strike a perfect balance with the best in racing. \\
In a week a Bengals team will again be the man to be the ninth and third best in the tournament to earn a league victory. \\
This team won and will be a big help to the boys XXL team as they record their best game of the year. \\
I bet they have your best goal of the day already. \\
   \midrule
   AE + RTT \\ 
   \midrule
                 They have to give it a try next season. \\
                Whoa, D’ghetto We have been the victors. \\
     You guys have managed to be the winner of this one. \\
 They even had a very big buck of success in this fight. \\
                   We've finished on the top one though. \\
                       Away you go the champions, thief. \\
                  We have the best victory in the races. \\
            Have you encountered the biggest finish yet? \\
               This makes your second bracket excellent. \\
                         We have won the flagship event! \\
    \bottomrule
  \end{tabular}
}
\end{table}

\pagebreak
\section{Full results for the noise experiments}
\label{app:noise-full-experiments}
The full results on the compact noise space $\mathcal{O}$ are shown in Figure~\ref{fig:alpha-noise-length}.
\begin{figure}[htb!]
\centering
\caption{Full quantitative results on the compact set $\mathcal{O}$ that contains different values of noise around the embedding.}
\label{fig:alpha-noise-length}
\subfloat[]{%
  \includegraphics[clip,width=0.85\linewidth]{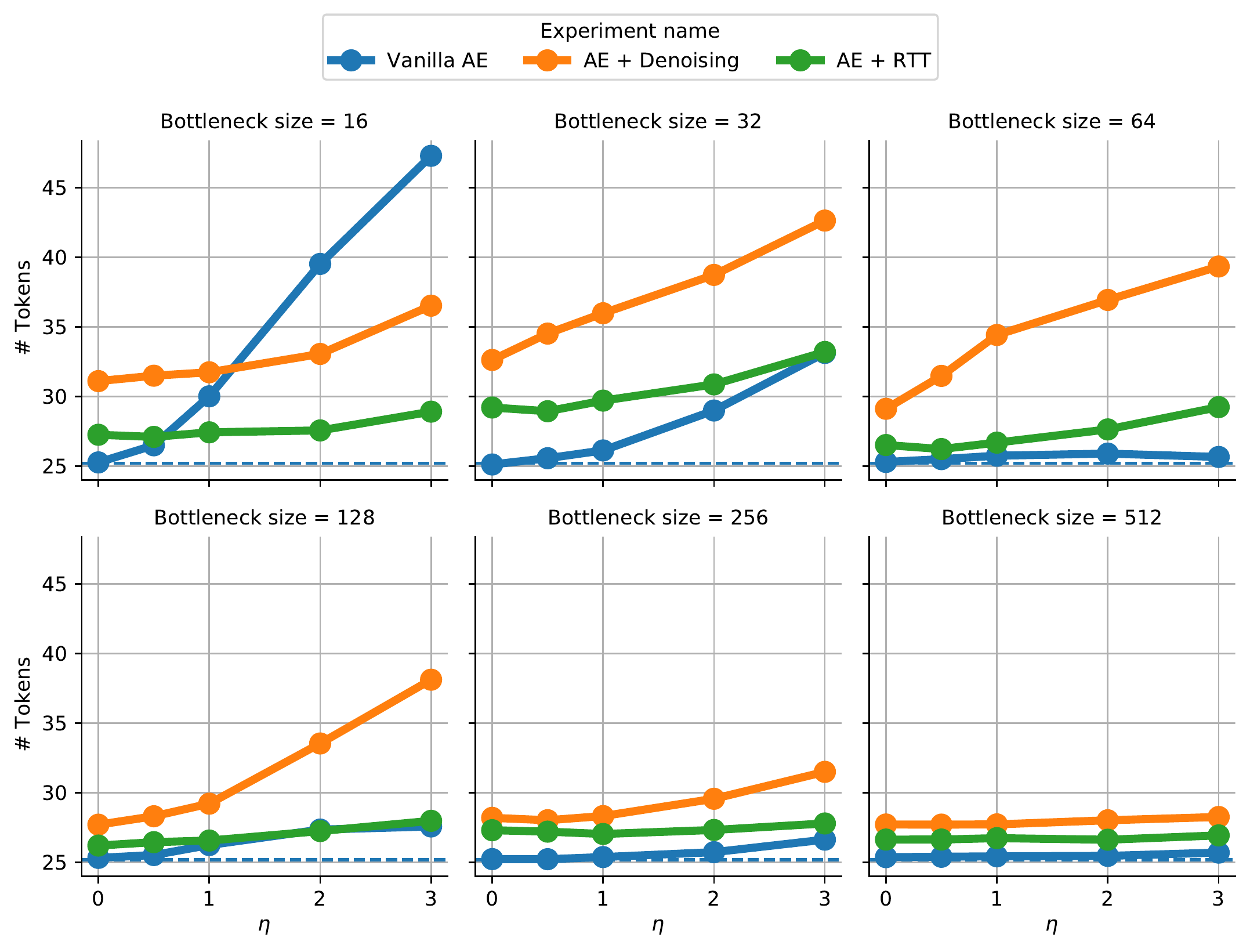}%
}
\vspace{-0.6cm}
\subfloat[]{%
  \includegraphics[clip,width=0.85\linewidth]{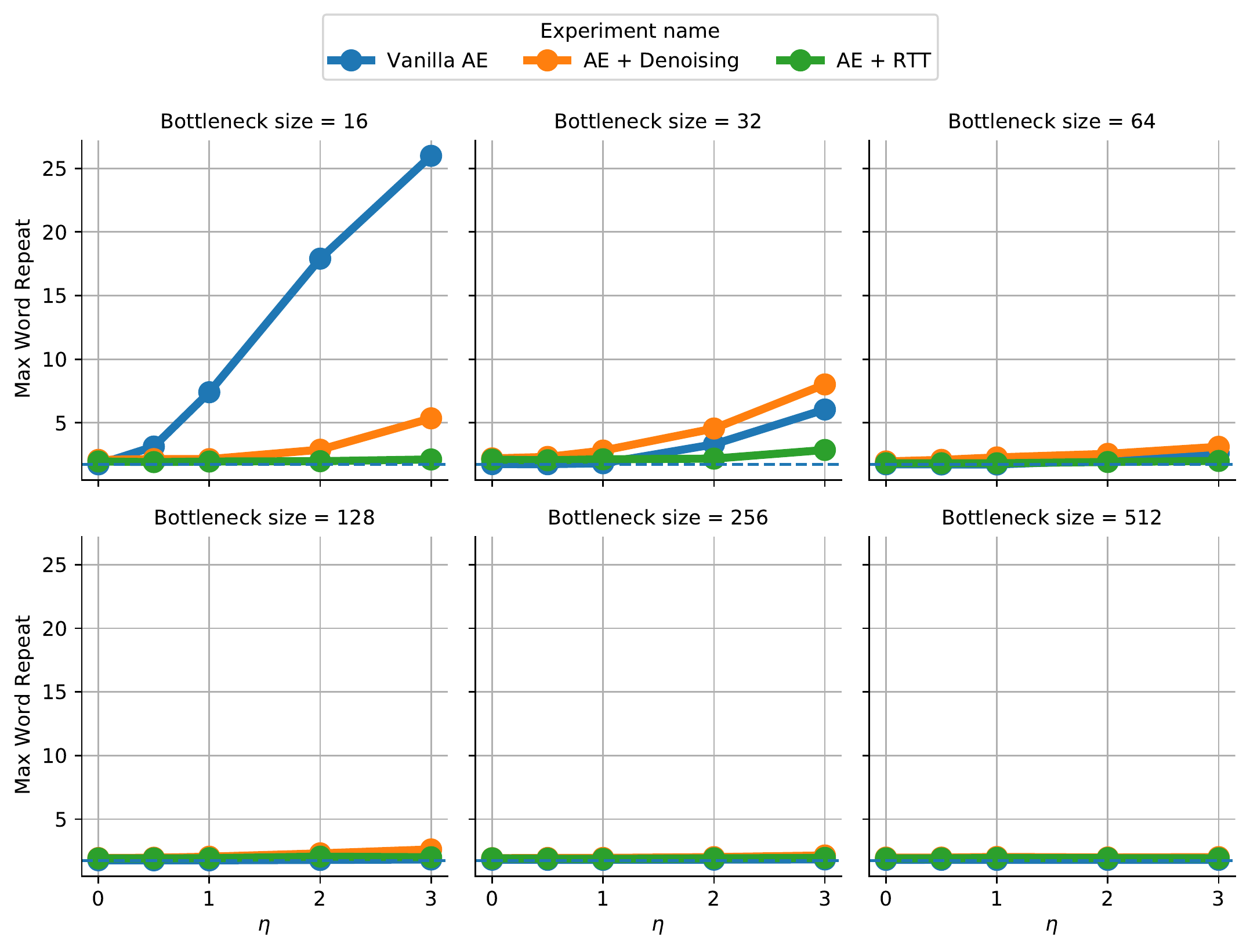}%
}
\end{figure}

\begin{figure}[h]\ContinuedFloat
\centering
\subfloat[]{%
  \includegraphics[clip,width=0.90\linewidth]{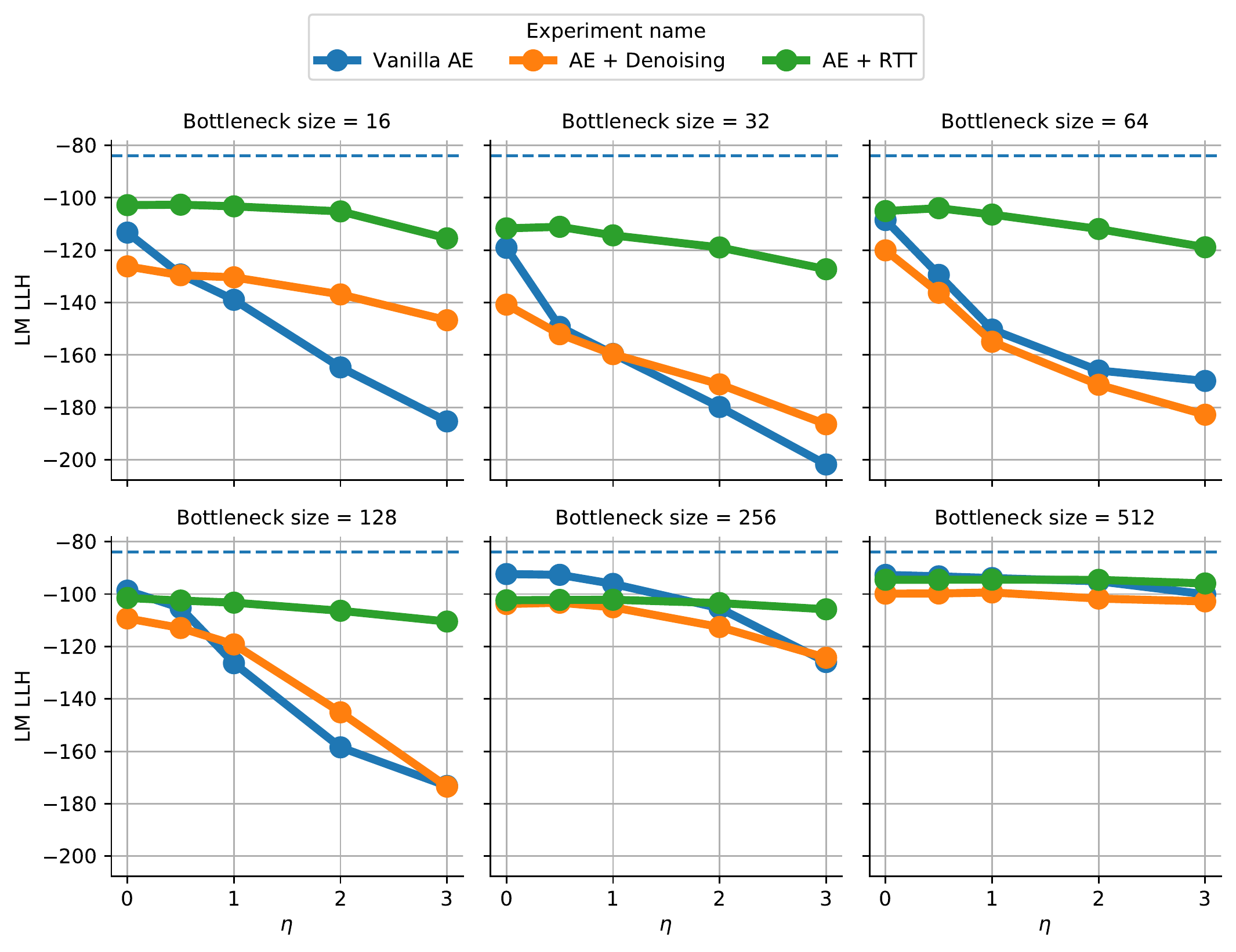}%
}

\subfloat[]{%
  \includegraphics[clip,width=0.90\linewidth]{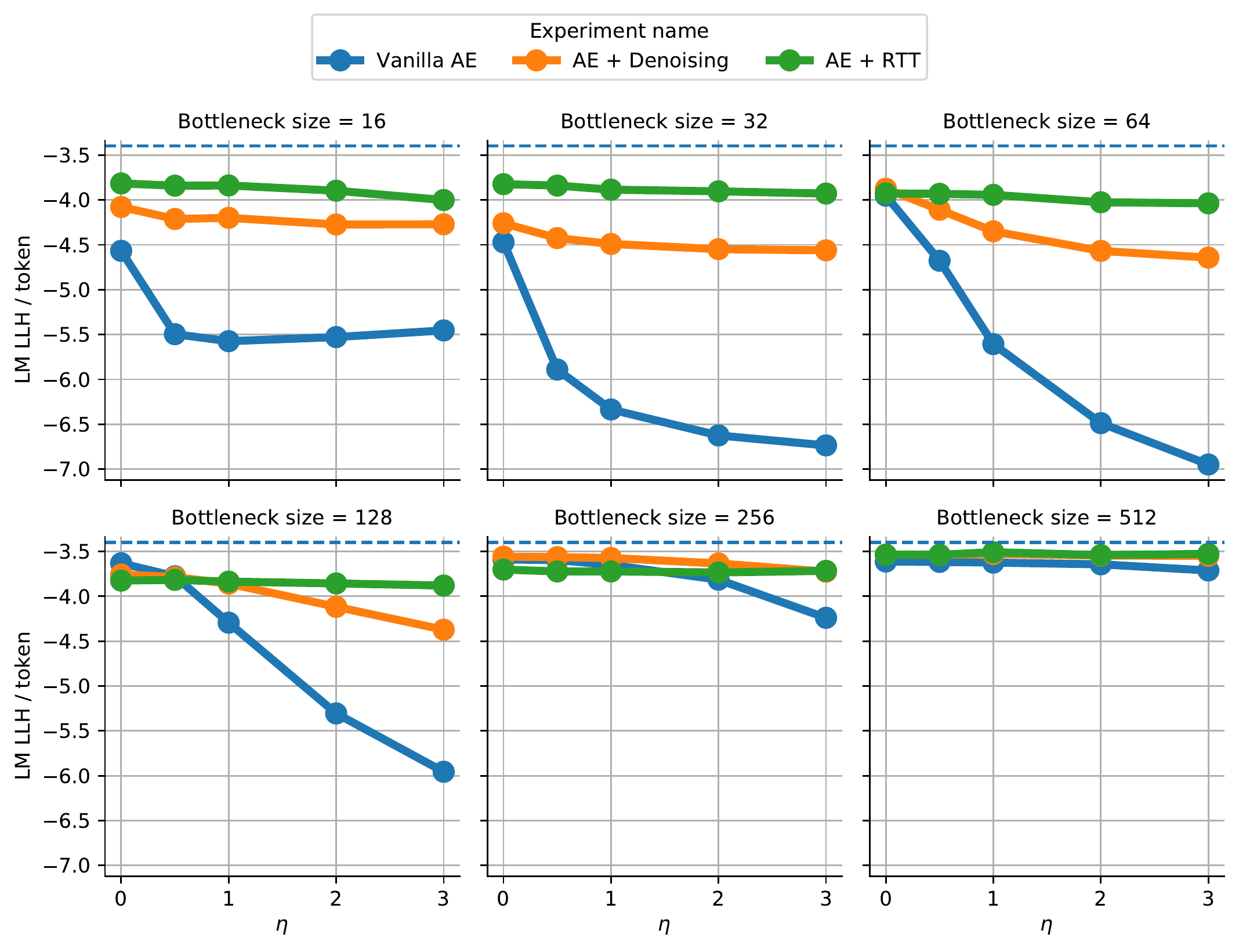}%
}
\label{fig:alpha-noise-lm}
\phantomcaption
\end{figure}

\begin{figure}[h]\ContinuedFloat
\centering
\subfloat[]{%
  \includegraphics[clip,width=0.9\linewidth]{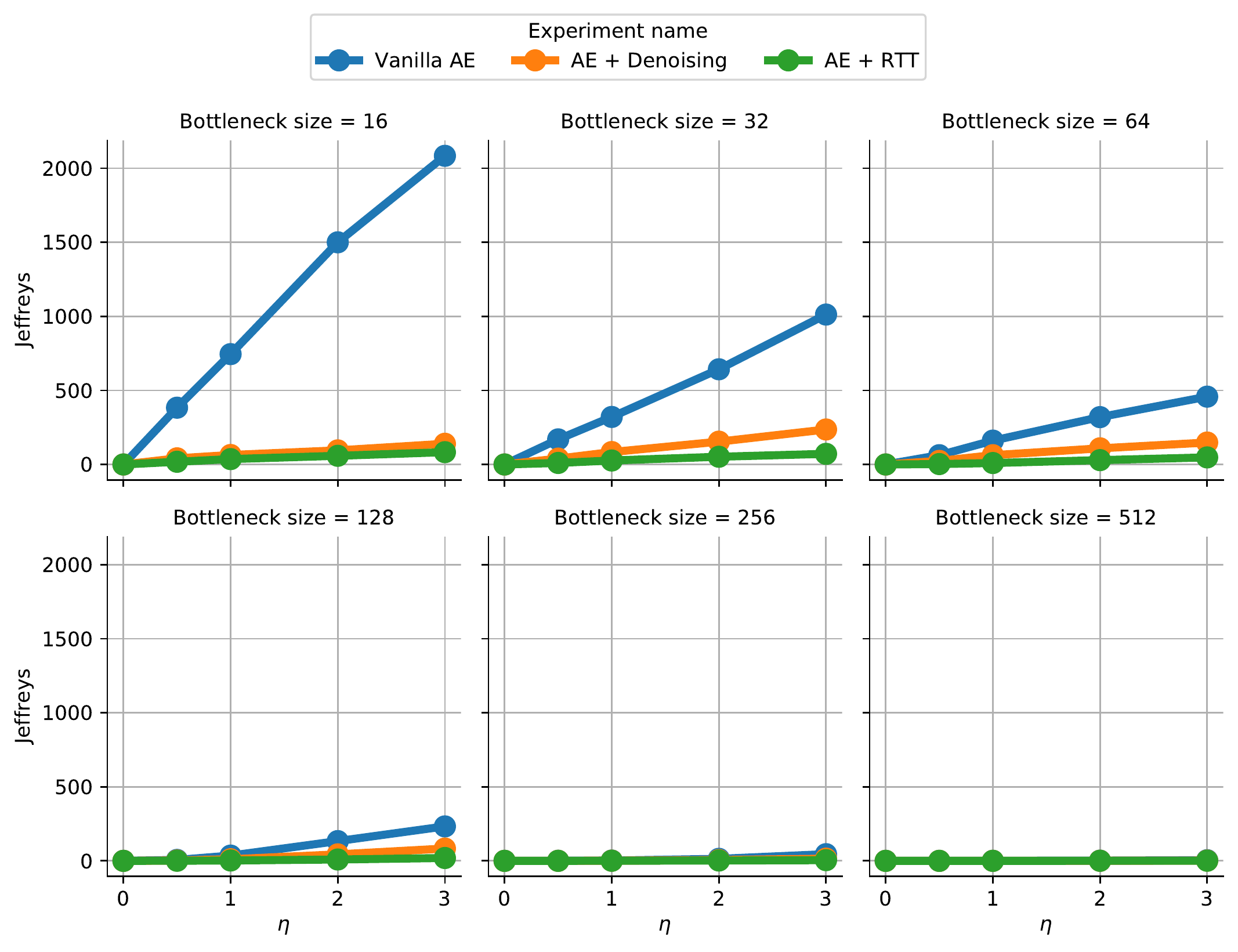}%
}
\label{fig:alpha-noise-jeffreys}
\phantomcaption
\end{figure}

\begin{figure}[h]\ContinuedFloat
\centering
\subfloat[]{%
  \includegraphics[clip,width=0.90\linewidth]{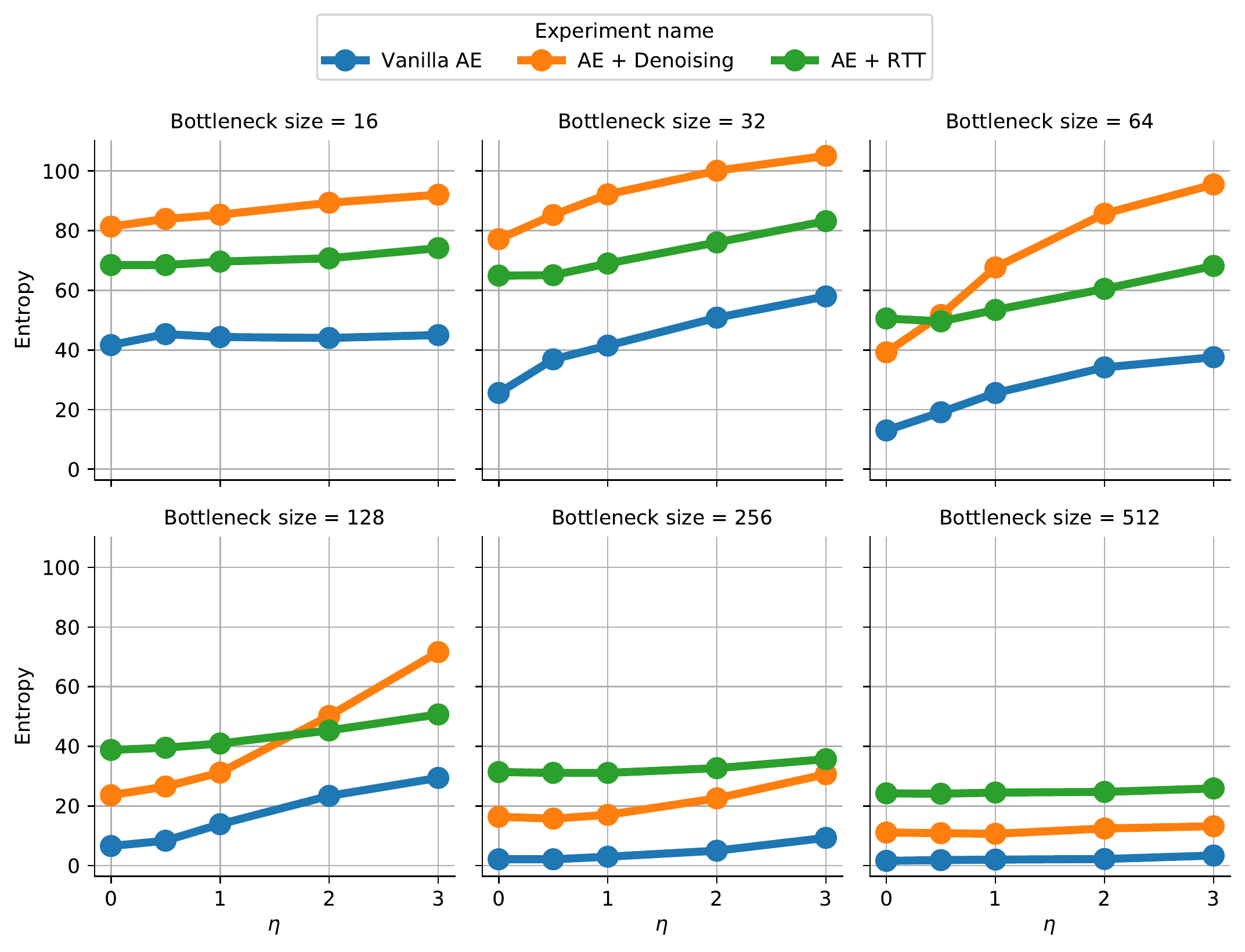}%
}

\subfloat[]{%
  \includegraphics[clip,width=0.90\linewidth]{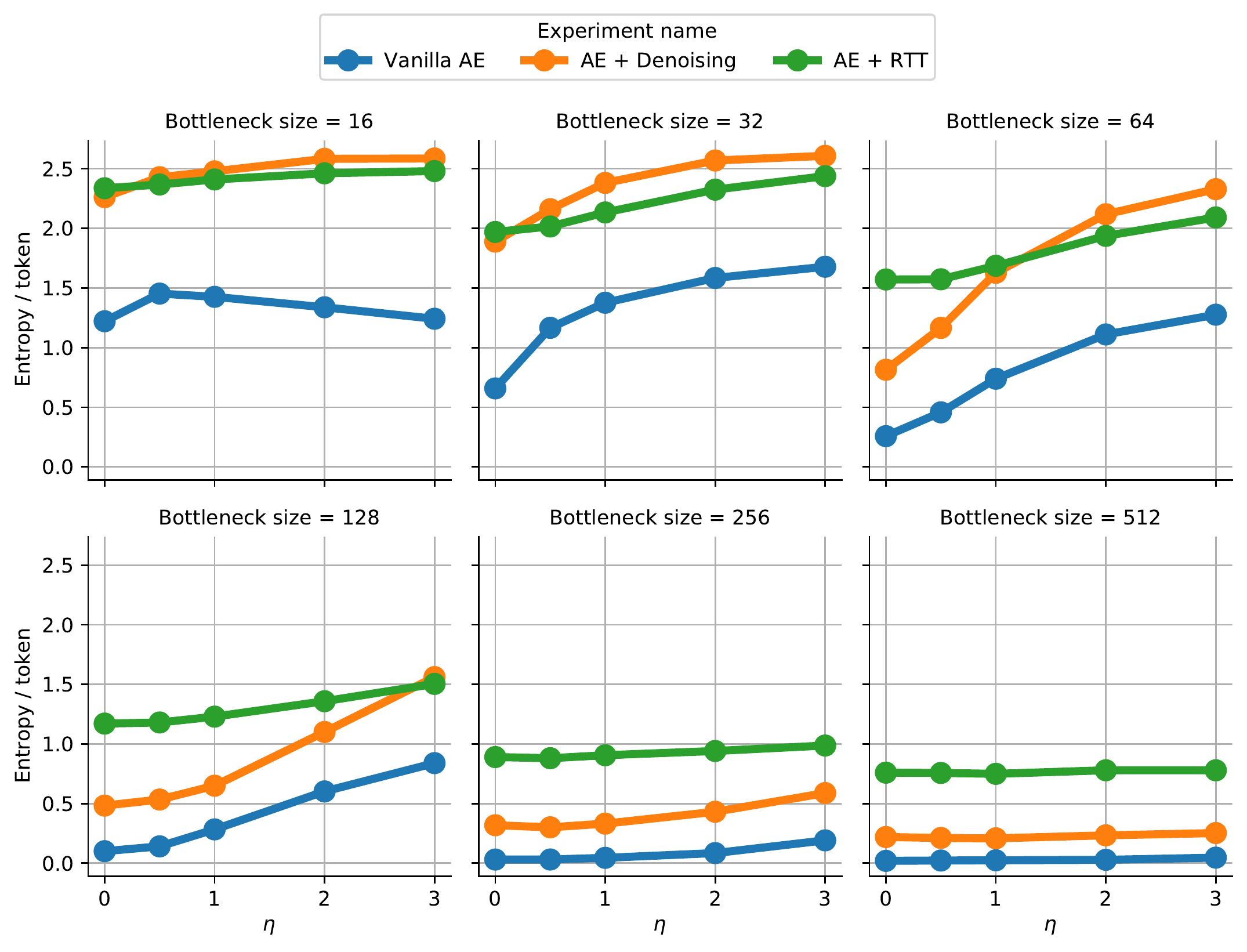}%
}
\label{fig:alpha-noise-entropy}
\phantomcaption
\end{figure}

\end{document}